\documentclass[10pt]{article}
\usepackage[accepted]{style}

\usepackage{amssymb}
\usepackage{amsthm}
\usepackage{bm}
\usepackage{mathtools}
\usepackage{parskip}
\usepackage{graphicx}
\usepackage{enumitem}
\usepackage{csquotes}
\usepackage{url}
\usepackage{booktabs}
\usepackage[colorlinks=true, allcolors=gray!80!black]{hyperref}
\usepackage{algorithm}
\usepackage{algorithmic}
\usepackage{cleveref}
\usepackage{tikz-cd}
\usepackage[many,listings]{tcolorbox}
\usepackage{float}
\usepackage{listings}
\usepackage{pythonhighlight}
\usepackage{mathrsfs}
\usepackage{listings}
\usepackage{xcolor}

\newcommand{\R}{\mathbb{R}}
\newcommand{\N}{\mathbb{N}}
\newcommand{\eps}{\varepsilon}
\newcommand{\inp}[1]{{#1}_{\mathcal{F}}} \newcommand{\inpc}{c} 
\newcommand{\inpd}{d} 
\newcommand{\out}[1]{{#1}_{\mathcal{G}}} \newcommand{\outc}{C} 
\newcommand{\outd}{D} 
\newcommand{\K}{K}
\DeclareMathOperator*\infp{\vphantom{p}inf}

\theoremstyle{definition}

\colorlet{pygray}{black!40!white}
\colorlet{pylightgray}{black!40!white}
\definecolor{pyblue}{RGB}{126,166,224}
\definecolor{pyred}{RGB}{234,107,102}

\lstdefinestyle{pythonstyle}{
    language=Python,
    basicstyle=\ttfamily\small,
    keywordstyle=\bfseries\color{black},
    stringstyle=\color{black},
    commentstyle=\color{pygray}\itshape,
    identifierstyle=\color{black},
    morekeywords=[1]{def},
    keywordstyle=[1]{\color{teal}\bfseries},
    morekeywords=[2]{if,else,for,while,return,import,from,as,with,try,except,finally,class,True,False,None,and,or,not,in,is,lambda,yield,break,continue,pass,range,int,float,list,dict,set,str,enumerate,zip,len},
    keywordstyle=[2]{\color{teal}\bfseries},
    morekeywords=[3]{Parameter,randn,rfft,einsum,cfloat,pad,irfft,unsqueeze,sum,cat,view,cdist,expand,arange,sin,cos,backward,step,square,abs,det,stack,column_stack,pi,parameters},
    keywordstyle=[3]{\color{pyred}},
    morekeywords=[4]{MLP,spectral_conv,pointwise_layer,integral_transform,encoder_decoder,NeuralOperator,Adam,train_iteration,Delauney,compute_weights},
    keywordstyle=[4]{\color{pyblue}},
    emph=[5]{Tensor},
    emphstyle=[5]{\color{pylightgray}},
    moredelim=**[s][\color{pylightgray}]{Tensor[}{]},
    deletekeywords=[2]{sum,abs},
    showstringspaces=false,
    breaklines=true,
    tabsize=4,
    keepspaces=true,
    columns=flexible,
    xleftmargin=4pt,
    xrightmargin=4pt,
    framexleftmargin=12pt,
    framexrightmargin=12pt,
    upquote=true,
    morestring=[b]',
    morestring=[b]"
}

\renewcommand{\thealgorithm}{\arabic{algorithm}}
\newtcblisting{mypython}[2][]{
    enhanced,
    float,
    floatplacement=t,
    colback=black!5!white,
    colframe=black!15!white,
    arc=5pt,
    boxrule=0pt,
    fonttitle=\small\color{black},
    listing only,
    listing options={style=pythonstyle},
    before title={\refstepcounter{algorithm}\label{#1}},
    title={Algorithm \thealgorithm: #2},
    top=-2pt,
    bottom=-2pt,
    left=2pt,
    right=2pt,
}

\newcommand{\tldr}[1]{
\begin{tcolorbox}%
[enhanced,colback=black!5!white,colframe=black!5!white,arc=5pt,boxrule=0pt]

  #1
\end{tcolorbox}
}

\setlist[itemize]{leftmargin=*}
\setlist[enumerate]{leftmargin=*}
\lstdefinestyle{float}{
  float=tp,
  floatplacement=tbp,
}

\crefname{lstlisting}{algorithm}{algorithm}
\Crefname{lstlisting}{Algorithm}{Algorithm}

\title{Principled Approaches for Extending Neural Architectures \\ to Function Spaces for Operator Learning}

\author{\name{Julius Berner}\affilnum{1}\thanks{Equal contribution}\phantom{$^*$}, \name{Miguel Liu-Schiaffini}\affilnum{2}\footnotemark[1]\phantom{$^*$}, \name{Jean Kossaifi}\affilnum{1}, \name{Valentin Duruisseaux}\affilnum{2}, \name{Boris Bonev}\affilnum{1}, \name{Kamyar Azizzadenesheli}\affilnum{1}, \name{Anima Anandkumar}\affilnum{2} \\
\affilnum{1}\addr{NVIDIA} \affilnum{2}\addr{Caltech}}

\begin{document}

\maketitle

\begin{abstract}
A wide range of scientific problems, such as those described by continuous-time dynamical systems and partial differential equations (PDEs), are naturally formulated on function spaces. While function spaces are typically infinite-dimensional, deep learning has predominantly advanced through applications in computer vision and natural language processing that focus on mappings between finite-dimensional spaces. Such fundamental disparities in the nature of the data have limited neural networks from achieving a comparable level of success in scientific applications as seen in other fields. \emph{Neural operators} are a principled way to generalize neural networks to mappings between function spaces, offering a pathway to replicate deep learning's transformative impact on scientific problems. For instance, neural operators can learn solution operators for entire classes of PDEs, e.g., physical systems with different boundary conditions, coefficient functions, and geometries. A key factor in deep learning's success has been the careful engineering of neural architectures through extensive empirical testing. Translating these neural architectures into neural operators allows operator learning to enjoy these same empirical optimizations. However, prior neural operator architectures have often been introduced as standalone models, not directly derived as extensions of existing neural network architectures. In this paper, we identify and distill the key principles for constructing practical implementations of mappings between infinite-dimensional function spaces. Using these principles, we propose a recipe for converting several popular neural architectures into neural operators with minimal modifications. This paper aims to guide practitioners through this process and details the steps to make neural operators work in practice\footnote{Our code is available at \href{https://github.com/neuraloperator/NNs-to-NOs}{\nolinkurl{github.com/neuraloperator/NNs-to-NOs}}.}.
\end{abstract}

\section{Introduction to Neural Operators}
\label{sec:intro}

\tldr{
In many scientific and engineering applications, data is derived from an underlying function, of which we have access to discretizations at different resolutions. We want to develop deep learning models that \textbf{respect the functional nature of the data} in the sense that they are \textbf{agnostic to the discretization and resolution}.
}

In the past decade, deep learning has led to unprecedented advances in domains like computer vision, speech, and natural language processing. Following this success, deep learning is poised to have an arguably even greater transformational impact on the natural sciences. While underlying data in all these areas comes in vastly different structures, the standard models powering deep learning, neural networks, consume data in a unified way: processed versions in the form of finite-dimensional \emph{vectors}. For instance, these can be pixel representations of images and videos, word embeddings of language, or measurements of physical systems.

Yet, many physical phenomena of interest are inherently captured by \emph{continuum} descriptions.
Physical quantities describing, e.g., fluid and thermodynamics, continuum mechanics, or electromagnetism, depend on continuous variables such as spatial and temporal coordinates. Mathematically, such quantities are given as \emph{functions} depending on continuous variables, and their behavior is often governed by differential equations modeling the relationship between physical quantities and their derivatives.

For instance, we can describe the temporal evolution of a given physical system, such as a heat distribution in a certain domain. This naturally leads to the task of predicting the physical system after a given time (i.e., the solution \emph{function}) based on an initial condition of the system (i.e., an initial \emph{function}). Common tasks in computer vision are inherently defined on functions as well since the underlying natural scenes or objects depend on spatiotemporal variables. Images or videos merely capture a discretized version with a certain resolution; e.g., an image is a discretization of a two-dimensional function.

While classical neural networks can process discretizations of continuous quantities (e.g., on a mesh or grid), they either rely on a fixed discretization or do not generalize to unseen resolutions. However, we would like the action of the neural network to be invariant\footnote{The term \enquote{invariance} is a useful analogy; however, strictly speaking, neural operators still incur a (controllable) discretization error. Thus, we often say that a neural operator is \enquote{agnostic} to the discretization and produces \enquote{consistent} outputs for different resolutions. We introduce the appropriate mathematical concept of \enquote{discretization convergence} in \Cref{sec:operator_learning}.} to the specific choice of discretization and its resolution. Furthermore, in numerous applications, it is essential for the output to be a function that can be queried at arbitrary coordinates, allowing for further operations (e.g., differentiation and integration). To this end, neural networks have recently been generalized to so-called \emph{neural operators}. Neural operators are naturally formulated to operate on \emph{functions} rather than \emph{vectors}. In particular, \emph{by construction}, they output functions that can be queried at arbitrary coordinates and that are consistent across different discretizations of the input function; more precisely the outputs only differ by a discretization error that vanishes as the discretization is refined. This property of being agnostic to the underlying discretization enables learning of function-to-function mappings and multi-scale phenomena for the wide range of tasks where the underlying data is captured by continuum descriptions. This has led to significantly improved performance and generalization, as well as powerful capabilities, such as zero-shot super-resolution, in a series of practical applications. 
In turn, these advances have spurred a large interest in developing different neural operator architectures and training paradigms. The increased popularity of neural operators has also led to the development of dedicated Python libraries, such as the \textsc{NeuralOperator} library~\citep{kossaifi2024neural}, as well as large-scale benchmarks of these methods~\citep{ohana2024well}.

Using neural networks to represent continuous functions, known as \emph{implicit neural representations} or \emph{neural fields}, has a long history~\citep{lagaris1998artificial}. 
Special cases include \emph{physics-informed neural networks} (\textsc{PINN}s)~\citep{raissi2019physics}, where the neural network approximates the solution function of a partial differential equation (PDE) by optimizing its parameters to reduce deviations from the governing equation. Shortly after the introduction of neural operators, 3D scenes in graphics and computer vision have been similarly represented using \emph{neural radiance fields} (\textsc{NeRF})~\citep{mildenhall2020nerf,sitzmann2020implicit,jeong2022perfception}\footnote{It is worth noting that the recent architectures~\citep{shukla2021parallel,muller2022instant} share close similarities to early neural operator architectures such as (multipole) graph neural operators~\citep{li2020multipole}.}. In both of these cases, the neural network represents a \emph{single} function, i.e., a single PDE solution or 3D scene. In contrast, neural operators can be viewed as generalizations of neural fields because they output a function for any given input function; instead of representing a single function, they can learn operators that act on functions. In other words, neural operators can be viewed as \emph{conditional neural fields}~\citep{azizzadenesheli2024neural}, conditioned on different input functions. 
Alternatively, one can condition the implicit neural representations on the parameters of a (given or learned) finite-dimensional parametrization of the space of input functions~\citep{chen2022crom, serrano2023operator}. While this can be viewed as a special case of an encoder-decoder neural operator (see~\Cref{sec:enc_dec}), neural operators are more general and can handle input and output functions based on point evaluations without the need for parametrizations.

\paragraph{Contributions} In this work, we identify and distill the core principles for constructing neural operators as practical and principled instantiations of parametrized mappings between function spaces. Using these principles, we then propose a recipe for converting popular neural architectures into neural operators with minimal modifications. This paper aims to guide practitioners through this process and details the steps to design and apply neural operators successfully in practice.

\begin{enumerate}
    \item We \textbf{identify and motivate the main design principles of neural operators} and how they are necessary for well-posed operator learning (\Cref{sec:operator_learning}). In particular, neural operators are mappings between function spaces parametrized by a finite number of parameters. They should be discretization-agnostic and be able to approximate any sufficiently regular operator with arbitrarily low error. 

    \item Using the design principles of neural operators, \textbf{we establish a recipe for naturally extending successful architectures and layers to function spaces}, which we develop throughout \Cref{sec:nns_to_nos} and summarize in \Cref{sec: summary conversion}. We also demonstrate throughout \Cref{sec:nns_to_nos} that many popular neural operator architectures can be constructed using this procedure. In doing so, we also categorize existing neural operator architectures and layers. More specifically, we show that
    \begin{itemize}
        \item \textbf{Multilayer perceptrons} (\Cref{sec:fnn}) can be converted to integral transforms by parametrizing their weights and biases with learnable functions.
        \item \textbf{Convolutional neural networks} (\Cref{sec:cnn}) can be converted to convolutional integral operators by parametrizing the kernel as a learnable function.
        \item \textbf{Graph neural networks} (\Cref{sec:gnn}) become discretization-agnostic graph neural operators when quadrature weights are used for aggregation and neighborhoods are defined using subsets of the domain.
        \item \textbf{Self-attention in transformers} (\Cref{sec:transformer}) can be turned into resolution-agnostic self-attention using a similar strategy as for graph neural networks.
        \item \textbf{Encoder-decoder neural network architectures} (\Cref{sec:enc_dec}) can be generalized to function spaces by using parametric function classes for both the encoder and decoder.
    \end{itemize}
    We also show how other key building blocks of neural operator architectures (e.g., pointwise operators, positional encodings, and normalizations) can be similarly constructed from their finite-dimensional counterparts.
    
    \item We outline important \textbf{practical considerations for training neural operators} in \Cref{sec:training_main}. While many best practices from deep learning also directly apply to neural operators, we discuss in more detail the use of data and physics losses on function spaces (\Cref{sec:losses}), and data augmentation, optimization, and auto-regressive modeling in the context of operator learning (\Cref{sec:training}).
    
    \item We survey the diverse landscape of neural operator architectures that have been designed and used successfully across a wide range of \textbf{scientific and engineering applications} (\Cref{sec:applications_of_nos}).

    \item Lastly, we \textbf{perform empirical experiments} (\Cref{sec:experiments}) to illustrate the importance of certain design choices for neural operators, such as fixing the receptive field with respect to the underlying domain and using quadrature weights for aggregations. We also analyze the effects of multi-resolution training and different interpolation strategies.
\end{enumerate}

The more technical mathematical details are purposefully deferred to the appendix, to enhance the clarity of the presentation and prioritize an intuitive understanding. We refer the reader to~\cite{kovachki2023neural,boulle2023mathematical,kovachki2024operator,subedi2025operator} for a more theoretical treatment.

\section{From Discrete to Continuous: Learning on Function Spaces}
\label{sec:operator_learning}

\begin{figure}[t]
    \centering
    \includegraphics[width=\textwidth]{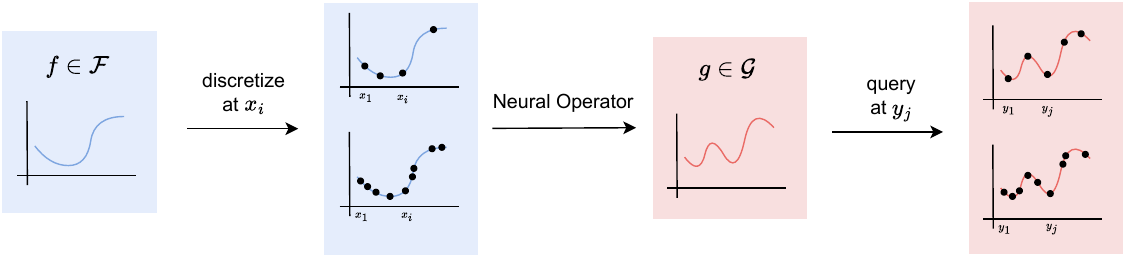}
    \vspace{-1.5em}
    \caption{\textbf{Illustration of a neural operator.} The input is a function $f\in\mathcal{F}$ that can be given at different discretizations $(x_i)_{i=1}^n$. The output is a function $g\in \mathcal{G}$ that can be queried at different points $(y_j)_{j=1}^m$.}
    \label{fig:no}
\end{figure}

Classical neural networks aim to approximate functions 
\begin{equation}
    x \longmapsto f(x)
\end{equation}
between (finite-dimensional) Euclidean spaces, such as $\R^d$. More precisely, they take a \emph{vector} $x$ as input and output another \emph{vector} $f(x)$. In contrast, we want to construct mappings, so-called \emph{operators}, 
\begin{equation}
\label{eq:operator}
  f  \longmapsto g
\end{equation}
between (infinite-dimensional) function spaces\footnote{We use $\mathcal{F}$ and $\mathcal{G}$ to denote the function spaces for the input and output functions, which are typically suitable subspaces of continuous functions.}, such as continuous functions. Operators take a \emph{function} $f$ 
as input and output another \emph{function}
$g$; see \Cref{tab:notation} for a comparison and \Cref{sec:notation} for details on our notation.

\begin{table}[t]
    \centering    
    \caption{Notation and comparison of neural networks to neural operators.} \vspace{0.5em}
    \begin{tabular}{lccccc}
    \toprule
     & \textbf{Neural Network} & \textbf{Neural Operator} & \textbf{Input} & \textbf{Output}  \\
    \midrule
    \textbf{Space} & Euclidean spaces (e.g. $\mathbb{R}^d$) & Function spaces & $\mathcal{F}$ 
    & $\mathcal{G}$ \\
    \textbf{Elements} & Vectors & Functions & $f \in \mathcal{F}$ & $g \in \mathcal{G}$  \\
    \textbf{Domain} & Integer indices & Real-valued coordinates & $x\in D_f$ & $y\in D_g$ \\
    \bottomrule
    \end{tabular}
    \label{tab:notation}
\end{table}

The transition from discrete to continuous problems requires a change in how we represent and manipulate data. 
Naturally, while we are interested in learning mappings between functions, we typically do not have access to the actual functions numerically. Instead, neural operators are designed to work on a \emph{discretization} $((x_i,f(x_i)))_{i=1}^n$ of a function $f$, assuming that we observe the function on a set of points $x_i$ (i.e., a point cloud or grid; see \Cref{sec:fun_representation} for alternative representations). Moreover, the output of the neural operator, which is another function $g$, can be \emph{queried} at arbitrary coordinates $y$ in the domain of $g$; see~\Cref{fig:no} for an illustration.

\subsection{Defining Principles of Neural Operators}
\tldr{
Neural operators represent a principled extension of neural networks to function spaces that are  parametrized by a \textbf{finite number of learnable parameters}. They \textbf{universally approximate function-to-function mappings} and provide \textbf{consistent predictions across resolutions}.}

To learn a mapping between functions in practice, we would want to satisfy the following properties (see \Cref{sec:no} for details):
\begin{itemize}
\item \textbf{Discretization-agnostic:} the learned mapping should be applicable to functions given at any discretization and produce \emph{consistent} outputs across resolutions. 
\item \textbf{Fixed number of parameters}: the number of learnable parameters needs to be \emph{fixed} and independent of the discretization.
\item \textbf{Universal approximation}: the family of mappings should be able to (provably) approximate, with arbitrarily low error, any sufficiently regular operator. 
\end{itemize}

\emph{Neural Operators} were designed from first principles to satisfy these properties and offer a natural framework for learning mappings between function spaces. In practice, neural operators still operate on discretized functions, but also respect the underlying functions they model and do not overfit to a fixed set of observation points. In particular, their outputs are functions that can be evaluated on arbitrary discretization, and the outputs of a neural operator for different discretizations only differ by a discretization error that can be made arbitrarily\footnote{In practice, we want to design neural operators where the discretization error is sufficiently small for the resolutions considered in applications.} small by refining the discretization---we say that the neural operator is \emph{discretization convergent}; see~\Cref{sec:disc_conv}.

\subsection{Features of Neural Operators}
\tldr{
Neural operators can be applied to data at different discretizations and \textbf{generalize across resolutions} by design. This is more \textbf{data-efficient}, \textbf{prevents overfitting} on the training resolution, and allows to \textbf{predict at arbitrary resolutions}.
}

Neural operators provide a principled framework to learn mappings on functional data. In the following we mention some of the unique features and advantages which are unlocked by neural operators and their ability to train and test on data with varying discretizations:

\begin{itemize}
    \item \textbf{Well-posedness:} For many tasks, the ground truth mapping is naturally an operator, i.e., a mapping between \emph{functions}. If we train a neural network that is not discretization-agnostic, we can overfit to the training discretization(s) (see \Cref{sec:gnn}). Operator learning allows us to approximate the ground truth operator and establish rigorous theoretical guarantees (see \Cref{app:error}). In addition, having output functions that can be queried arbitrarily is useful for downstream applications that require derived quantities such as derivatives or integrals. 
    \item \textbf{Data efficiency:} Standard numerical solvers often create their meshes adaptively, which naturally leads to datasets with different discretizations~\citep{li2024geometry,li2023fourier,codano}. Due to high computational costs, only a few high-resolution solutions are available in many applications, while a larger number of approximate solutions can be computed on discretizations with lower resolutions. Operator learning enables learning on different discretizations and resolutions simultaneously, ensuring sufficient data for training while maintaining the ability to resolve the underlying physics from the data at higher resolution. In practice, neural operators can work directly with functions discretized on point clouds with an arbitrary number of points and do not need an explicit mesh or partition of the domain into connected cells or elements.

    \item \textbf{Curriculum learning and faster training:} Neural operators can be trained via curriculum learning, where simpler, low-resolution samples are first used for training, gradually progressing to more challenging, high-resolution samples. 
    Such an adaptively refinement of the discretization during training can not only improve performance but also accelerate convergence and reduce the computational cost compared to training only on high-resolution data~\citep{george2022incremental,lanthaler2024discretization}.

    \item \textbf{Flexible inference:} A trained neural operator can be queried at arbitrary resolution. For instance, we can obtain consistent predictions for discretizations that have not been in the training data, both at lower as well as higher resolutions, i.e., zero-shot super-resolution~\citep{kovachki2023neural}.
\end{itemize}

Motivated by the properties and capabilities of neural operators, the next natural question is how to construct and implement them in practice. This will be covered in the next section, in particular outlining which adaptations existing neural network layers require.

\section{From Networks to Operators: Building Blocks of Neural Operators}
\label{sec:nns_to_nos}

\begin{figure}[t]
    \centering
    \includegraphics[width=0.9\textwidth]{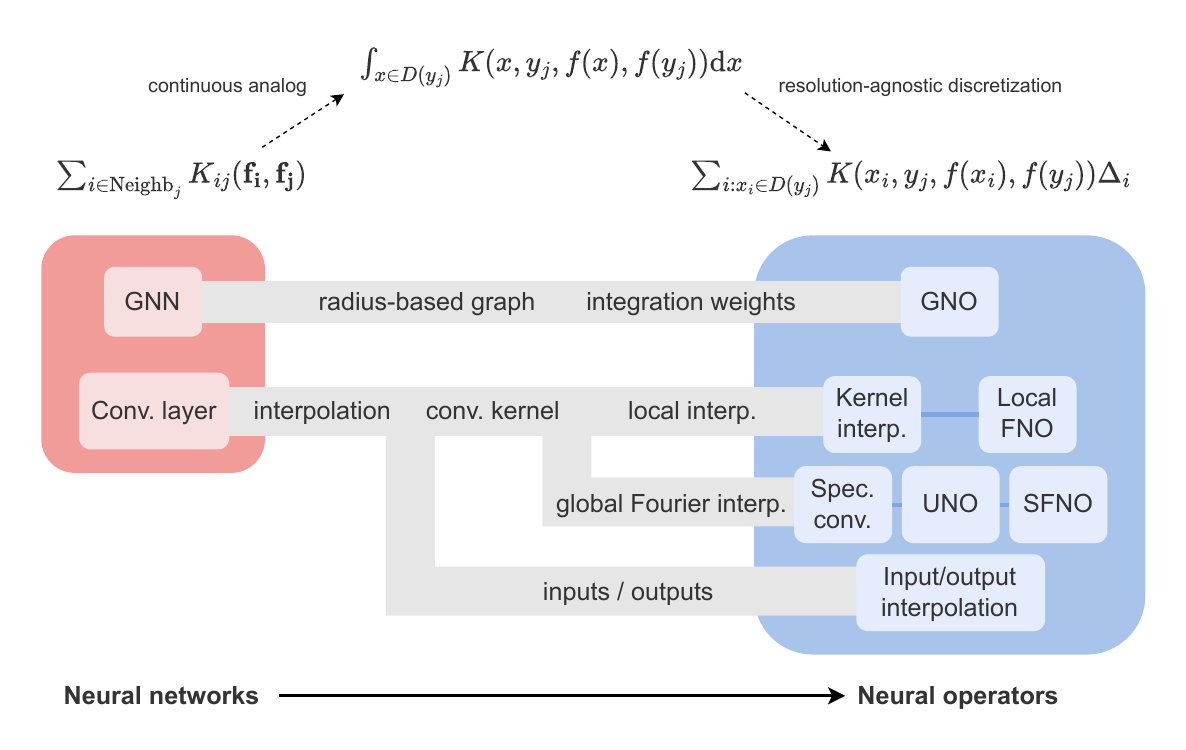}
    \caption{\textbf{Pipeline of converting neural networks to neural operators.} Graph neural network (GNN) and convolutional layers can be converted into well-posed neural operator layers through a sequence of simple modifications. \textsc{GNO} refers to the graph neural operator~\citep{li2020neural}, \enquote{Spec. conv.} refers to a spectral convolution as used in Fourier neural operators (\textsc{FNO}s)~\citep{li2020fourier}, and \textsc{Local FNO} refers to a \textsc{FNO} supplemented with local integral kernels~\citep{liu2024neural}. We denote by \textsc{UNO} the U-shaped neural operator~\citep{rahman2022u} and by \textsc{SFNO} the spherical Fourier neural operator~\citep{bonev2023spherical}. The overall strategy to convert neural networks into neural operators is outlined in \Cref{sec: summary conversion}.}  
    \label{fig:layers}
\end{figure}

Following a common paradigm in deep learning, an effective approach to designing neural operators is to construct simple and modular layers that can be composed to form expressive neural operator models. Each layer is itself an operator such that the intermediate outputs after each layer are \emph{functions} (as opposed to \emph{vectors} in standard neural networks). In particular, the discretized version of the neural operator can have a different discretization in each layer. In this section, we present how the most popular neural network layers can be converted to neural operator layers, ensuring they stay consistent across resolutions; see~\Cref{fig:layers} for a summary.

\subsection{From Fully-Connected Layers to Integral Transforms}
\label{sec:fnn}
\tldr{
  Integral transforms can be obtained by \textbf{parametrizing the weights and biases} of a fully-connected neural network layer \textbf{with learnable functions}. Instead of directly learning the weights for a fixed set of indices, we learn a function that maps input and query points to the value of the weights \textbf{weighted by the density of the input points}:
  \begin{equation}\nonumber
      \bm{g}_j = \sum_i \bm{\K}_{ji} \bm{f}_i + \bm{b}_j \text{\quad becomes \quad} g(y_j) = \sum_{i} \K(x_i,y_j) f(x_i) \Delta_i + b(y_j).
  \end{equation}
}

We start by looking at the most basic building block of multi-layer perceptions (\textsc{MLP}s) and neural networks: the fully-connected layer. \textsc{MLP}s can simply be described by a learnable affine-linear map (we discuss the non-linearity in~\Cref{sec:pointwise}), and their parameters are typically learned end-to-end via backpropagation. However, the affine-linear map assumes a fixed, finite-dimensional vector as input. In this section, we present a natural extension to affine-linear operators on infinite-dimensional function spaces. 

\paragraph{Fully-connected layer}
Assume we are given $n$ values $\bm{f}_i=f(x_i)$ of 
a function $f$ at points $(x_i)_{i=1}^n$ and a fully-connected layer with weight matrix $\bm{\K} \in \R^{m \times n}$ and bias vector $\bm{b} \in \R^{m}$. The $j$-th coordinate of the output of the fully-connected layer can then be written as 
\begin{equation}   
\label{eq:fnn}
\bm{g}_{j} \coloneqq (\bm{\K}\bm{f} + \bm{b})_j = \sum_{i=1}^n \bm{\K}_{ji}\bm{f}_i + \bm{b}_j.
\end{equation}
The weights are learned for \emph{fixed input and output dimensions} $n$ and $m$, which does not allow to work with different discretizations. 

\begin{figure}[t]
    \centering
    \includegraphics[width=\linewidth,clip,trim={77pt 25pt 0 0}]{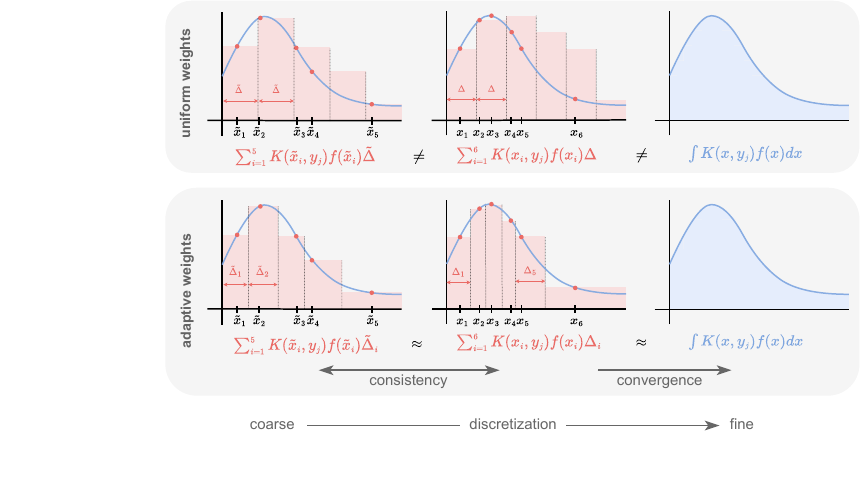}  \vspace{-3mm} \caption{\textbf{Visualizing the need for quadrature weights}. Aggregating function values at irregularly-spaced points without proper quadrature weights (e.g., taking the mean, i.e., $\Delta=1/n$, as in the top figure) adds more weight to densely sampled areas. When increasing the resolution, the output depends on the chosen refinement of the discretization and does not have a unique limit (\textbf{top}). The use of quadrature weights leads to consistent approximations for different discretizations that converge to a unique integral as the discretization is refined (\textbf{bottom}).}
    \label{fig:quadrature}
\end{figure}

\paragraph{Integral transform} To make this operation agnostic of the discretization, we assume that $\bm{\K}_{ji}$ and $\bm{b}_{j}$ are evaluations of \emph{functions} $\K$ and $b$ at input points $x_i$ and query points $y_j$. The result is a mapping from the input \emph{function} $f$ to an output \emph{function} $g$ given by
\begin{equation}
\label{eq:fnn_fun}
    g(y_j) \coloneqq  \sum_{i=1}^n \K(x_i, y_j)f(x_i) \Delta_i +b(y_j) \approx \int_{D_f} \K(x, y_j)f(x) \, \mathrm{d}x + b(y_j),
\end{equation}
where we formally defined\footnote{Depending on the codomains of $f$ and $g$, the output of $K$ can be a scalar or a matrix. In the latter case, diagonal matrices or low-rank representations are often chosen for efficiency~\cite{kovachki2023neural}.} $ \bm{\K}_{ji} = \K(x_i, y_j)\Delta_i $ and $\bm{b}_j = b(y_j)$ in~\eqref{eq:fnn}. Specifically, note that we introduced so-called \emph{quadrature weights} $\Delta_i$, which ensure that the sum approximates the integral representation and that the output is agnostic to the resolution. For instance, we can view the sum as a \emph{Riemann sum} for the integral, which converges as we refine the discretization. For one-dimensional grids, we can take $\Delta_i=\frac{x_{i+1} - x_{i-1}}{2}$ (with suitable definitions at the boundary); see \Cref{fig:quadrature}. For more general point clouds, we can partition the domain $D_f$ and take the volume of the subregion containing $x_i$; see~\Cref{sec:quadrature} and~\Cref{alg:quad}. The \emph{kernel} and \emph{bias functions} $K$ and $b$ are typically parametrized by neural networks and learned end-to-end using backpropagation.

\begin{mypython}[alg:quad]{Quadrature weights from a Delaunay triangulation in $d=2$ (see~\Cref{sec:fnn} and~\Cref{sec:quadrature})}
def compute_weights(x: Tensor[n_in, 2]):
    tri = Delaunay(x)  # Compute Delaunay triangulation, e.g., from scipy.spatial
    delta = zeros(len(x)) # Initialize the quadrature weights
    for simplex in tri.simplices:
        points = x[simplex] # Vertices of the simplex (i.e., triangle in d=2)
        volume = 0.5 * abs(det(column_stack((points[1]-points[0], points[2]-points[0]))))
        for idx in simplex:
            delta[idx] += volume / 3 
    return delta
\end{mypython}

Unlike the standard definition in~\eqref{eq:fnn}, the expression in~\eqref{eq:fnn_fun} can now be evaluated for an arbitrary number of input and output points $n$ and $m$. This naturally extends the definition of a fully-connected layer, an affine-linear mapping between vectors $\bm{f}$ and $\bm{g}$, to an affine-linear operator between functions $f$ and $g$. Crucially, as we increase the resolution (i.e., refine the discretization), the model converges to the unique \emph{integral operator} in~\eqref{eq:fnn_fun}, an instantiation of a \emph{graph neural operator} (\textsc{GNO})~\citep{li2020neural}; see also ~\Cref{sec:gnn} and~\Cref{alg:gno}.

\subsection{From Elementwise Functions to Pointwise Operators} 
\label{sec:pointwise}

\tldr{
Operations that are applied pointwise to the discretized function are \textbf{naturally agnostic to the resolution}. This includes activation functions as well as liftings and projections. However, they can be \textbf{queried only at the input points}.}

We did not cover activation functions of the fully-connected layers in \Cref{sec:fnn}. However, we will see that this concept transfers directly to neural operators as part of the broader class of pointwise operators.

\paragraph{Elementwise function}
Several common layers in deep learning apply the same operation to each element of the input vector, such as activation functions, $(1\times 1)$-convolutions, or certain normalization layers. They can directly be used in neural operators if these elements correspond to the pointwise evaluations of the input function $\bm{f}=(f(x_i))_{i=1}^n$. In other words, there is a (potentially learnable) function $K$ such that the layer can be written as
\begin{equation}
    \bm{g}_i = K(\bm{f}_i).
\end{equation}

\paragraph{Pointwise operator} 
We can then define a pointwise operator, also called \emph{Nemytskii operator}~\citep{renardy2006introduction}, mapping a function $f$ to a function $g$, by\footnote{We note that $K$ can also depend on both $f(x_i)$ and $x_i$. This is included in the definition by assuming that the coordinates are included in the output of the function, which is typically done in the form of a positional encoding; see~\Cref{sec:aux}.}
\begin{equation}
    g(x_i)=K(f(x_i)).
\end{equation}
Such mappings are agnostic to the resolution as they can be applied to any discretization of the input function~$f$. However, the output function $g$ can only be queried at the same points $x_i$ where the input function $f$ is given, i.e., the discretizations of the functions $f$ and $g$ are necessarily the same. To query $g$ at arbitrary points, one can compose the pointwise operator with another layer (e.g., in the simplest case, an interpolation layer; see also~\Cref{sec:aux}).  

Pointwise operators allow transferring all activation functions from neural networks to neural operators. Moreover, they can be used for skip-connections and so-called lifting and projection layers. Specifically, given an input function $f\colon D_f \to \R^c$ with codomain $\R^c$, we can choose $K\colon \R^c \to \R^C$ (e.g., as a neural network) to map to a function $g\colon D_g \to \R^C$ with codomain $\R^C$, i.e., where the function is \enquote{lifted} $(C > c)$ or \enquote{projected} $(C < c)$; see~\Cref{alg:spectral_conv}.
Inspired by image processing, one also refers to $c$ as the number of \emph{channels} since a discretization $\bm{f}$ on $n$ points has shape\footnote{Depending on the order of the dimensions, such layers can also be efficiently realized using $(1\times 1)$-convolutions.} $n \times c$. If the number of channels is adaptive (e.g., for multiphysics simulations with different physical variables), one can use attention mechanisms for the function $K$~\citep{codano}.

\subsection{From Convolutional Neural Networks to Spectral Convolutional Operators}
\label{sec:cnn}
\tldr{For discrete local convolutions, the receptive field is tied to the resolution and becomes increasingly local as the resolution increases. By \textbf{parametrizing the kernel as a learnable function}, convolutional operators can be obtained as special cases of integral operators \textbf{with a fixed receptive field}:
  \begin{equation}\nonumber
      \bm{g}_j = \sum_{i = j - k}^{j+k}  \bm{\K}_{j-i} \bm{f}_{i}  \text{\quad becomes \quad} g(y_j) = \Delta \!\!\!\!\! \sum_{i\colon |x_i - y_j| \le r } \!\!\!\!\! \K(y_j - x_i) f(x_i).
  \end{equation}
For \textbf{global convolutions}, this can be efficiently realized \textbf{using spectral convolutions}, leading to Fourier neural operators.}

A core building block of modern neural network architectures is the convolutional layer. It can be viewed as a special case of a linear layer as in~\eqref{eq:fnn} and, similarly, can be extended to a special case of the integral operator in~\eqref{eq:fnn_fun}. Note that we omit the bias term for simplicity since it can be handled as in the previous subsection.

\begin{figure}[t]
    \centering
     \includegraphics[width=0.85\textwidth]{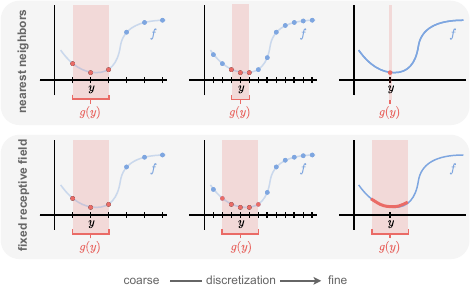}
    \caption{\textbf{Illustration of collapsing receptive fields with a nearest neighbors strategy.} The figure shows the values of the input function $f$ (blue) which influence the output function $g$ at a point $y$ when using a nearest neighbors strategy (e.g. as in convolutional and graph \emph{neural networks}) (\textbf{top}) and with a fixed receptive field (e.g. as in convolutional  and graph \emph{neural operators}) (\textbf{bottom}). If the neighborhood is selected using a nearest neighbors strategy, the receptive field (red) collapses when the discretization is refined (from \textbf{left} to \textbf{right}).}
    \label{fig:receptive_field}
\end{figure}

\paragraph{Convolutional layer} We will see that the receptive field of convolutional layers changes when directly applying their discrete formulation to different resolutions. To demonstrate this, recall that the output of a one-dimensional convolutional\footnote{We treat the one-dimensional case only for simplicity, and the arguments naturally extend to higher dimensions. Moreover, we note that most modern deep learning frameworks such as PyTorch and JAX perform a \emph{cross-correlation}, which differs from the convolution by a reflection of the kernel, i.e., a sign flip. This does not impact our presentation since the kernel is learned.} layer with kernel $\bm{\K}=(\bm{K}_i)_{i=-k}^k \in \R^{2k+1}$ of size $2k+1$ can be written as
\begin{equation}
\label{eq:conv_layer}
\bm{g}_j = (\bm{\K} \ast \bm{f})_j = \sum_{i = j - k}^{j+k}  \bm{\K}_{j-i} \bm{f}_{i},
\end{equation}
where $\bm{f}_{i}=f(x_i)$ are $n$ evaluations of the input function $f$ on an \emph{equidistant} grid $x_i \coloneqq i\Delta$ of width $\Delta$ (with suitable padding). In principle, such a convolution can be applied to discretizations of $f$ on grids of any width $\Delta$. 
However, this operation becomes increasingly local (in other words, the receptive field of the convolution shrinks) when decreasing the width $\Delta$ or, equivalently, increasing the number of discretization points. Specifically, notice that the output $\bm{g}_j$ in~\eqref{eq:conv_layer} depends on $f$ only at an interval of size $2k\Delta$ including the points $x_{j-k}$ to $x_{j+k}$. 
When increasing the resolution, i.e., in the limit $\Delta \to 0$, the convolution thus converges to the pointwise operation\footnote{Strictly speaking, this would converge to a pointwise operator as described in~\Cref{sec:pointwise}, however, with a very inefficient parametrization.} $\bm{g}_j = f(x_j) \sum_{i}  \bm{\K}_{i}$ (by continuity of $f$); see~\Cref{fig:receptive_field}. Equivalently, this can be understood as zero-padding the kernel to the number of discretization points.

\paragraph{Convolutional operator}
    To make the receptive field independent of the underlying resolutions, we can leverage the integral operator from~\Cref{sec:fnn} and again view $\bm{\K}_{j-i}$ as the evaluation of a (univariate) function $K$ at the point $y_j - x_i$. Given that $K$ has support in $[-r,r]$, we can then define the corresponding convolutional neural operator
\begin{equation}
\label{eq:cnn_fun}
 g(y_j) = \Delta \!\!\!\!\! \sum_{i\colon |x_i - y_j| \le r} \!\!\!\!\! \K(y_j - x_i) f(x_i) \approx \int_{y_j-r}^{y_j+r} \K(y_j - x)f(x) \, \mathrm{d}x = (K \ast f)(y_j),
\end{equation}
mapping the input function $f$ to an output function $g$. Here, we formally defined $ \bm{\K}_{j-i} = K(y_j - x_i)  \Delta$ with $k = \frac{r}{\Delta}$ in~\eqref{eq:conv_layer}, where $\Delta$ are the quadrature weights for the equidistant\footnote{Using non-uniform quadrature weights $(\Delta_i)_{i=1}^n$ as in~\Cref{sec:fnn}, it is straightforward to generalize the convolutional operator to general point clouds $(x_i)_{i=1}^n$.} grid. In particular, the receptive field in~\eqref{eq:cnn_fun} is now always of size $2r$, independent of the discretization of $f$. Moreover, the output function $g$ can be queried at arbitrary points $y_i$ (not necessarily grid points) since the function $K$ in~\eqref{eq:cnn_fun} can be evaluated at any point in its domain.

We can again parametrize $K$ by neural networks or other parametric families of functions\footnote{Similar approaches can be found in the context of \emph{hypernetworks}, \emph{scale-equivariant}, and \emph{discrete-continuous (group) convolutions}; see~\cite{liu2024neural} for an overview.}. One possibility is to interpolate the discrete kernel $\bm{K}$ and scale by the quadrature weights, which also works for a pre-trained layer (see~\Cref{sec:experiments}).
We note that, due to the fixed receptive field, the computational cost of the convolutional operator in~\eqref{eq:cnn_fun} scales as $\mathcal{O}(rn^2)$ (where $\Delta \sim \frac{1}{n}$) instead of $\mathcal{O}(kn)$ for the convolutional layer in~\eqref{eq:conv_layer}.

If $f$ and $K$ are periodic, one possibility to obtain (quasi-)linear cost in $n$ is to leverage the \emph{convolution theorem} and rewrite the convolution in~\eqref{eq:cnn_fun} as a multiplication in Fourier space, i.e.,
\begin{equation}
    K \ast f = \mathscr{F}^{-1} \left( \mathscr{F}(K) \mathscr{F}(f) \right),
\end{equation}
where $\mathscr{F}$ denotes the mapping from a periodic function to its Fourier series. Instead of parametrizing the function $K$, we can now parametrize a \emph{fixed} number of modes, i.e., Fourier coefficients $\mathscr{F}(K)$ (implying that $K$ has full support). This is often referred to as a \emph{spectral convolution} and represents the main building block of the \emph{Fourier Neural Operator} (\textsc{FNO})~\citep{li2020fourier}; see~\Cref{alg:spectral_conv} and \Cref{sec:fno} for details.

\begin{mypython}[alg:spectral_conv]{Pointwise operator from~\Cref{sec:pointwise}, and spectral convolution from~\Cref{sec:cnn} in $d=1$.}
net = MLP(dim_in=codim_in, dim_out=codim_out) # MLP parametrizing the pointwise layer

def pointwise_layer(f: Tensor[bs, n_in, codim_in]) -> Tensor[bs, n_in, codim_out]:
    return net(f)

# Parameters for spectral convolution
params = Parameter(randn(modes, codim_in, codim_out, dtype=cfloat) * init_std)

def spectral_conv(f: Tensor[bs, n_in, codim_in], n_out=None) -> Tensor[bs, n_out, codim_out]:
    f_ft = rfft(f, dim=1)
    g_ft = einsum("bmi,mio->bmo", f_ft[:, :modes, :], params) # Multiply modes with parameters
    return irfft(g_ft, n=n_out or n_in, dim=1) 
\end{mypython}

In practice, $\mathscr{F}$ and its inverse can be approximated\footnote{We refer to~\citet{bartolucci2023neural,lanthaler2024discretization} for an analysis of the discretization error (also referred to as \emph{aliasing} error) depending on the grid resolution and spectrum of the input function.} with the \emph{Discrete Fourier Transform} and accelerated using the \emph{Fast Fourier Transform} (for sufficiently large grid resolutions and number of modes~\citep{lingsch2023beyond}) leading to an efficient implementation of~\eqref{eq:cnn_fun}. This can also be viewed as a Fourier interpolation of the kernel $\bm{K}$ in~\eqref{eq:conv_layer} to other resolutions. While such global convolutions are typically not used for tasks in computer vision, they can be important in scientific applications. 

Multiple techniques, such as incremental learning or tensorization of $\mathscr{F}(K)$, have been developed to reduce the computational costs~\citep{george2022incremental,kossaifi2023multi}. Moreover, \cite{rahman2022u} proposed \emph{U-shaped neural operators} (\textsc{UNO}) which reduce and increase the resolution of the discretization (and the number of modes) in deeper layers of the model (analogous to \textsc{U-Net}s~\citep{ronneberger2015u}). 

An alternative approach to interpolating the kernel, would be to interpolate the input function $f$ to match the training resolution~\citep{raonic2024convolutional}. However, this has the downside of discarding information beyond the training resolution. Yet another approach is to rescale and center the kernel of discrete convolutions in order to converge to differential operators~\citep{liu2024neural}.

\subsection{From Graph Neural Networks to Graph Neural Operators}
\label{sec:gnn}

\tldr{While graph neural networks can be natively applied to different discretizations of the input function, they do not necessarily generalize to unseen discretizations. To make them agnostic to the resolution, we need to \textbf{use quadrature weights for the aggregation} and \textbf{define neighborhoods using subsets of the domain}, independently of the discretization:
  \begin{equation}\nonumber  
    \bm{g}_j = \sum_{i \in \mathrm{Neighb}_j} K_{ij}(\bm{f}_i , \bm{f}_j)\text{\quad becomes \quad} g(y_j) = \sum_{i\colon x_i \in D(y_j)} \!\!\!\! \K(x_i,y_j,f(x_i),f(y_j)) \Delta_i.
  \end{equation}
}

In this section, we look at discretizations of functions on more general point clouds and typical layers of graph neural networks. In their standard form, these layers are not agnostic to the resolution and require the use of quadrature weights and fixed receptive fields, as seen in previous sections. This will lead us to \emph{Graph Neural Operators} (\textsc{GNO}s)~\citep{li2020neural}, a general form of (nonlinear) integral operators. Similar to how graph neural networks generalize feed-forward and convolutional neural networks, \textsc{GNO}s can be seen as a unifying framework for the neural operator layers discussed in~\Cref{sec:fnn,sec:cnn}.

\paragraph{Graph neural networks} 
Let us interpret our discretization $(x_i)_{i=1}^n$ as a graph with a given connectivity and denote by $\textrm{Neighb}_j$ the indices of points connected to $x_j$. The output of a prototypical \emph{message-passing} layer would then be defined as\footnote{We omit node-wise operations since they can be treated as described in~\Cref{sec:pointwise}. Moreover, we implicitly allow for edge features by letting $K$ depend on $i$ and $j$ and we assume that $j$ is contained in its own neighborhood $\textrm{Neighb}_j$.}
\begin{equation}
\label{eq:gnn}
   \bm{g}_j = \sum_{i \in \mathrm{Neighb}_j} K_{ij}(\bm{f}_i, \bm{f}_j),
\end{equation}
where $K$ is a neural network and the sum could also be replaced by other permutation-invariant aggregation functions, such as a (weighted) mean or maximum. We note that this definition also includes \emph{graph attention} and \emph{graph convolution} layers~\citep{bronstein2021geometric}. In particular, for nearest neighbor graphs on regular grids and $K_{ij}(\bm{f}_i, \bm{f}_j)=\bm{\K}_{j-i} \bm{f}_{i}$, we recover the convolutional layers discussed in~\Cref{sec:cnn}. 

\paragraph{Graph neural operators} As was done in the previous sections, we generalize~\eqref{eq:gnn} to the neural operator setting by: 
\begin{itemize}
    \item using the corresponding points $x_i$ (and hence explicit positional information) instead of indices $i$,
    \item aggregating the contributions in each neighborhood using quadrature weights $\Delta_i$, and
    \item making the receptive field independent of the discretization by defining the neighborhoods based on suitable subsets $D(y_j)$ of the underlying domain (e.g., a common choice are \emph{radius graphs} $D(y_j)=B_r(y_j)$, where points in a ball of certain radius $r$ around $y_j$ belong to the neighborhood).
\end{itemize}

In summary, this leads to the general (nonlinear) \emph{graph neural operator} (\textsc{GNO}) given by
\begin{equation}
\label{eq:gnn_fun}
g(y_j) = \sum_{i\colon x_i \in D(y_j)} \!\!\!\! \K(x_i,y_j,f(x),f(y_j)) \Delta_i \approx \int_{x \in D(y_j)} \!\!\!\! \K(x,y_j,f(x),f(y_j)) \, \mathrm{d}x,
\end{equation}
where we formally defined $\mathrm{Neighb}_j \coloneqq \{i \colon x_i \in D(y_j) \}$ and $\K_{ij} = \K(x_i,y_j,\cdot,\cdot)\Delta_i$ in~\eqref{eq:gnn}. It is easy to see that \textsc{GNO}s can represent both local and global integral operators depending on the choice of $D(y_j)$. In particular, the definition in~\eqref{eq:gnn_fun} subsumes\footnote{The operator in~\eqref{eq:fnn_fun} is a \textsc{GNO} with $\K(x,y_j,f(x),f(y_j)) = \K(x, y_j)f(x)$ and $D(y_j)=D_f$, while the operator in~\eqref{eq:cnn_fun} is a \textsc{GNO} where $\K(x,y_j,f(x),f(y_j)) =\K(y_j - x)f(x) $ with $D(y_j)=B_r(y_j)$.} and generalizes the integral operators defined in~\eqref{eq:fnn_fun} and~\eqref{eq:cnn_fun}. When using a local integral operator, \cite{lin2025mGNO} propose to multiply the kernel with a differentiable weight function supported on $D(y_j)$ to retain the differentiability of $g$ w.r.t.\@ the query points $y_j$ (which is needed for physics-based losses; see~\Cref{sec:losses}).

While we can use arbitrarily discretized functions $f$ as inputs to the \textsc{GNO}, the general formulation in~\eqref{eq:gnn_fun} only allows querying on the same points $y_j$.
To this end, one often uses kernels of the form $K(x,y_j,f(x))$ (or variants as in~\eqref{eq:fnn_fun} and~\eqref{eq:cnn_fun} leading to linear operators) that do not depend on $f(y_j)$ and allow for queries at arbitrary points~\citep{li2020neural,li2024geometry}; see~\Cref{alg:gno}. While the \textsc{GNO} can deal with arbitrary geometries and discretizations, the evaluation of the kernel and aggregation in each neighborhood can be slow and memory-intensive. For certain discretizations and domains, considering convolutional kernels as in~\Cref{sec:cnn}
allows to reduce the computational cost and establish equivariant\footnote{The approach in~\Cref{sec:cnn}, yields translation-equivariant layers for regular grids on the torus. We refer to~\citet{helwig2023group,xu2024equivariant,cheng2024equivariant,liu2024neural} for neural operators that are equivariant to other groups.} layers.  

\begin{mypython}[alg:gno]{Linear integral operator from~\Cref{sec:gnn} with radius graph and diagonal kernel.}
net = MLP(dim_in=2*dim, dim_out=codim) # MLP parametrizing the kernel

def integral_transform(
    f: Tensor[bs, n_in, codim], x: Tensor[bs, n_in, dim], delta: Tensor[bs, n_in],
    y: Tensor[bs, n_out, dim], radius=None
) -> Tensor[bs, n_out, codim]:
    # Kernel evaluation (if radius is not None, one should use sparse ops in CSR format)
    shape = [bs, n_in, n_out, dim]
    kernel_inp = [x.unsqueeze(2).expand(shape), y.unsqueeze(1).expand(shape)] 
    kernel = net(cat(kernel_inp, dim=-1)) 

    # Weighted aggregation using the quadrature weights
    kernel *= delta.view(bs, n_in, 1, 1)
    kernel *= f.unsqueeze(2)
    if radius is not None:
        kernel[cdist(x, y) > radius, :] = 0
    return kernel.sum(dim=1)
\end{mypython}

\subsection{From Transformers to Transformer Neural Operators}
\label{sec:transformer}

\tldr{
 Attention layers in transformers can be made discretization-agnostic \textbf{using quadrature weights for the normalization and aggregation}.
}

While transformers can be converted into neural operators using a similar strategy as the one for \textsc{GNN}s, we highlight specific considerations in this section. In particular, we will focus on the attention mechanism and refer to~\Cref{sec:pointwise} and~\Cref{sec:aux} for tokenwise operations and positional embeddings. 

\paragraph{Transformer}
 A typical self-attention mechanism can be written as
\begin{equation}
\label{eq:attention}
   \bm{g}_j  = \sum_{i=1}^n  \frac{K(\bm{f}_i, \bm{f}_j)}{\sum_{\ell=1}^n K(\bm{f}_\ell, \bm{f}_j)} v(\bm{f}_i),
\end{equation}
where $K$ is an attention mechanism, e.g., \emph{soft(arg)max with dot-product attention} $K(\bm{f}_i, \bm{f}_j)= \exp( \tau \langle k(\bm{f}_i), q(\bm{f}_j) \rangle)$ with temperature $\tau$ and learnable affine-linear mappings $k$, $q$, and $v$ (see~\Cref{sec:attention}). 

\paragraph{Transformer neural operator} The self-attention layer in~\eqref{eq:attention} can be viewed as a variant of a graph neural network layer as in~\eqref{eq:gnn} with fully-connected graph and an additional normalization (i.e., the denominator in~\eqref{eq:attention}). Due to the fully-connected graph, the receptive field is always the full domain, independent of the discretization. However, we still need to use quadrature weights for the aggregation and normalization to guarantee a unique limit when the discretization is refined. Doing so directly leads to the self-attention layer of \emph{transformer neural operators} given by
\begin{equation}
\label{eq:transformer_fun}
    g(y_j) = \sum_{i=1}^{n}  \frac{K(f(x_i), f(y_j))}{\sum_{\ell=1}^{n} K(f(x_\ell), f(y_j)) \Delta_\ell} v(f(x_i)) \Delta_i \approx \int_{D_f} \frac{ K(f(x), f(y_j))}{\int_{D_f}  K(f(z), f(y_j)) \, \mathrm{d}z} v(f(x)) \, \mathrm{d}x,
\end{equation}
which maps input functions $f$ to output functions $g$ and converges to an integral operator as the discretization is refined~\citep{kovachki2023neural}. We note that for equal quadrature weights $\Delta_i=\Delta$, e.g., for regular grids, the weights cancel, and we recover the standard attention mechanism in~\eqref{eq:attention}. Since the layer only depends on the values of $f$, we need to assume that they incorporate a positional embedding (e.g., by concatenating the coordinates $x_i$ to the output $f(x_i)$; see~\Cref{sec:aux}). Moreover, as explained in~\Cref{sec:gnn}, the input function $f$ can be discretized arbitrarily but $g$ can only be queried at the same discretization. To circumvent this limitation, one can use \emph{cross-attention} (see~\Cref{sec:attention}). 

Note that for self-attention layers, treating the points $f(x_i)$ as tokens leads to a quadratic cost in the size of the discretization, which might be prohibitive for fine resolutions. Alternatives include other variants of attention mechanism (without softmax and with
linear computational cost)~\citep{cao2021choose,li2022transformer,hao2023gnot,li2024scalable}, or tokenization along other dimensions (for instance, the channel dimension, i.e., dimension of the codomain of $f$~\citep{codano}; see also~\Cref{sec:pointwise}).

To reduce computational costs, vision transformers (e.g.\@ \textsc{V}i\textsc{T}s) use convolutional layers for the tokenization, i.e., they treat (non-overlapping) patches of the input as tokens~\citep{choromanski2020rethinking,vit}. However, these layers are dependent on the input discretization and need to be adapted as described in~\Cref{sec:cnn}. Similarly,~\citet{calvello2024continuum} define tokens by restricting the function to subsets of the domain (i.e., patches) and then use neural operators for tokenwise operations and inner products on function spaces for the attention mechanism. However, this requires sufficient resolution on each patch and can potentially create discontinuities at the boundaries of the patches.

\subsection{From Finite-Dimensional to Infinite-Dimensional Encoders \& Decoders}
\tldr{
We can construct neural operator layers based on encoders and decoders mapping between \textbf{finite-dimensional latent spaces} and function spaces. These are often based on parametric function classes where the encoder \textbf{infers the parameters} given the input function, a learned mapping \textbf{transforms the parameters in latent space}, and the decoder \textbf{outputs the corresponding function} specified by the transformed parameters.   
}

\label{sec:enc_dec}
A common technique in deep learning is to use encoders and decoders, i.e.,
\begin{equation} \label{eq:Decoder_K_Encoder}
    \bm{g} = \left(\operatorname{Decoder} \circ \, \K \circ \operatorname{Encoder}\right)(\bm{f}),
\end{equation}
such that we can operate with a learnable function $K$, typically a neural network, in a latent space. Such an idea can also be used to construct neural operator layers. However, for neural operators, we need to make sure that the encoder is agnostic to the discretization of the input function $f$ and that the output of the decoder can be queried at arbitrary resolution. In particular, we want to design 
encoders and decoders that map between function spaces and a finite-dimensional latent representations.
Different from neural networks, such encoder-decoder constructions are more often used for single layers of a neural operator and typically combined with pointwise skip-connections (see~\Cref{sec:pointwise}) to counteract the information bottleneck given by the finite-dimensional intermediate representation.

In the following, we present different practical versions of such encoders and decoders\footnote{For simplicity, we assume that $K\colon\R^k\to\R^k$ has the same input and output dimensions. We also note that the component functions of the input and output functions $f$ and $g$ are often considered separately for the encoder and decoder, but $K$ is applied to the combined latent representations.}. 

\paragraph{Encoder} Let us first describe different encoders, mapping discretizations of $f$ to a $k$-dimensional latent representation $\bm{v}=\operatorname{Encoder}(f)$, where $k$ does \emph{not} depend on the number of discretization points.
\begin{enumerate}
    \item \emph{Evaluation functionals:} In general, we can use $k$ functionals mapping from $\mathcal{F}$ to $\R$. An option used in \textsc{DeepONet}s~\citep{lu2019deeponet} are evaluation functionals, i.e., $\bm{v}_j = f(\xi_j)$, for fixed, so-called \emph{sensor points}~$\xi_j$. In this case, $\K$ is referred to as the \emph{branch net}. However, such an encoder ignores the values of $f$ at other points than $\xi_j$ and requires the sensor points to be a subset of the discretization points. 
    \item \emph{Inner products:} More generally, one can use approximations to inner products\footnote{If $\mathcal{F}$ is a Hilbert space, the \emph{Riesz representation theorem} shows that every continuous linear functional can be represented as an inner product. Note that for the $L^2$-inner product in~\eqref{eq:inner_product}, we denote by $[b_j(x_i)]^*$ the conjugate transpose of $b_j(x_i)$.} in function spaces, e.g.,
    \begin{equation}
    \label{eq:inner_product}
      \bm{v}_j = \sum_{i=1}^n [b_j(x_i)]^*  f(x_i) \, \Delta_i \approx  \int_{D_f} [b_j(x)]^* f(x) \, \mathrm{d}x  = \langle b_j, f\rangle_{L^2}
    \end{equation}
    with (potentially learnable) functions $b_j$ as proposed in~\textsc{DeepONet-Operator}s~\citep{kovachki2023neural}; see also~\Cref{alg:enc_dec}.

    \item \label{it:dict_enc} \emph{Projections}: We can project $f$ to the closest function in a family\footnote{To guarantee discretization convergence (see~\Cref{sec:disc_conv}), we require suitable assumptions on the function class and the optimization problem.} $f_{\bm{v}}$ parametrized by the vector $\bm{v}$, which is then used as the latent representation. For a general function class, e.g., neural networks\footnote{If $\bm{v}$ represents the weights of a neural network, $f_{\bm{v}}$ is often referred to as a \emph{neural field} or \emph{implicit neural representation}; see also~\Cref{sec:fun_representation}. For the function $K$, one can then use architectures that are specialized for operating on neural network parameters; see, e.g.,~\cite{zhou2023permutation,lim2024graph}.}, there is no closed-form expression for the optimal parameters $\bm{v}$, requiring an inner optimization loop~\citep{serrano2023operator}. 
    To this end, one often considers linear combinations $f_{\bm{v}} = \sum_{j=1}^{k} \bm{v}_j b_j$ given by a \emph{dictionary} of functions~$b_j$, e.g., subsets of classical approximation systems, such as polynomial bases, splines, Fourier bases, or wavelets~\citep{li2020fourier,gupta2021multiwavelet,tripura2022wavelet,bartolucci2023neural}. Finding the coefficients $\bm{v}$ then leads to a linear least squares problem and, for orthogonal systems, $\bm{v}$ can be inferred via inner products as in~\eqref{eq:inner_product}. For the Fourier basis, this corresponds to the discrete Fourier transform as used in the \textsc{FNO}.
\end{enumerate}

\paragraph{Decoder} Let us now present different decoders, mapping from a finite-dimensional vector $\bm{w}= K(\bm{v})$ to a function $g=\operatorname{Decoder}(\bm{w})$ that can be queried at different points $y$. 
\begin{enumerate}
    \item \label{it:dict_dec} \emph{Linear combinations:}
    The finite-dimensional vector $\bm{w}$ can be thought of as describing the coefficients of linear combinations\footnote{Mathematically speaking, such representations yield \emph{linear} approximation methods~\citep{devore1998nonlinear}.} $g_{\bm{w}} \coloneqq \sum_{j=1}^{k} \bm{w}_j b_j $, where, as before, $b_j$ are taken from a dictionary of functions. 
    The dictionary can also be adapted to a (training) dataset of functions. 
    For instance, one can use the first $k$ basis functions of an (empirical) \textsc{PCA} decomposition~\citep{bhattacharya2021model,lanthaler2023operator}.
    The functions $b_j$ can also be parametrized by neural networks (the so-called \emph{trunk net}) and learned end-to-end as is done in
    \textsc{DeepONet}s~\citep{lu2019deeponet}.

    \item \emph{Neural networks:} More generally, we can replace linear combinations of (learnable) dictionaries with a fully learnable neural network, where the input is given by the query point $y$ and the vector $\bm{w}$, as proposed in the \emph{Nonlinear Manifold Decoder} (\textsc{NOMAD})~\citep{seidman2022nomad}. Instead of using $\bm{w}$ as an input, we can also view it as (latent representation of the) parameters of a neural network~\citep{serrano2023operator}. In this case, the resulting encoder-decoder layer can be viewed as a meta-network outputting a \emph{neural field} conditioned on the input function $f$.
\end{enumerate}

\begin{mypython}[alg:enc_dec]{Encoder-decoder layer from~\Cref{sec:enc_dec} with inner product encoder and \textsc{\textsc{NOMAD}} decoder.}
encoder = MLP(dim_in=dim, dim_out=latent_dim*codim_in)
net = MLP(dim_in=latent_dim, dim_out=latent_dim)
decoder = MLP(dim_in=latent_dim+dim, dim_out=codim_out)

def encoder_decoder(
    f: Tensor[bs, n_in, codim_in], x: Tensor[bs, n_in, dim], 
    delta: Tensor[bs, n_in], y: Tensor[bs, n_out, dim]
) -> Tensor[bs, n_out, codim_out]:
    enc = encoder(x).view(bs, n_in, codim_in, latent_dim) * f.unsqueeze(-1)
    inner_products = (enc.sum(dim=2) * delta.unsqueeze(-1)).sum(dim=1)
    latent = net(inner_products)
    return decoder(cat([latent.unsqueeze(1).expand(bs, n_out, latent_dim), y], dim=-1))
\end{mypython}

In~\Cref{sec:special_enc_dec}, we show how certain integral operators, including spectral convolutions, can be viewed as encoder-decoder operators.
In particular, for spectral convolutions as employed in \textsc{FNO}s, we use a truncated Fourier basis for the inner products in the encoder and the linear combinations in the decoder. Similarly we can recover the integral operators used in the \emph{Laplace operator}~\citep{cao2023lno} or the \emph{(Multi-)Wavelet neural operator}~\citep{gupta2021multiwavelet,tripura2022wavelet} when replacing the Fourier basis by the (truncated) eigenbasis of the Laplacian and the Wavelet basis, respectively.

Finally, we note that encoder-decoder operators are universal for continuous operators, i.e., for any accuracy, there exist an encoder, a mapping $\K$, and a decoder approximating the operator uniformly on compact sets up to the given accuracy~\citep{kovachki2023neural} when composed as in~\eqref{eq:Decoder_K_Encoder}. However, in practice, we do not know the required size of the latent dimensions which governs the predefined parametric family of functions that the decoder can output. Thus, as mentioned before, one typically combines encoder-decoder layers with skip-connections to retain expressivity.

\subsection{Auxiliary Layers}
\label{sec:aux}

In this section, we outline some building blocks for neural operators that improve and stabilize performance or enable change of discretizations.

\paragraph{Positional encoding}
We can encode the \enquote{position}, i.e., the coordinates $x$, and add or concatenate it to the function values $f(x)$. The latter approach is often used as a first layer in neural operators to improve performance; see~\Cref{alg:train}. For the encoding of the position, often (non-learnable) sinusoidal embeddings as in \textsc{NeRF}s or transformers are used~\citep{mildenhall2020nerf,vaswani2017attention}.

In transformers, positional encodings are important since attention layers are permutation equivariant.
However, such encodings are often based on an absolute \emph{ordering} of the tokens. Since this does not generalize to our setting where tokens correspond to (arbitrary) discretizations of a function $f$, we concatenate or add embeddings of the absolute (or relative; see~\citet{li2022transformer,su2024roformer}) \emph{coordinates}. 

\paragraph{Normalization}
As for neural networks, normalization of the data as well as intermediate representations is important for training.
There are two important points to consider for the normalization of the data. First, we typically do not know the range of the data (e.g., in PDEs) and thus use \emph{standardization}, e.g., centering the data to mean zero and rescaling to unit variance. Further, for computing the mean and variance, it is important to take into account the functional nature of the data, e.g., we compute 
\begin{equation}
\label{eq:stats}
    \mu = \frac{\sum_{i=1}^n f(x_i) \Delta_i}{\sum_{i=1}^n \Delta_i} \approx \frac{1}{|D_f|} \int_{D_f} f(x) \, \mathrm{d}x,
\end{equation}
which takes into account quadrature weights $\Delta_i$ and is thus agnostic to the resolution. The standard deviation is calculated analogously.
We note that for equal weights, these statistics correspond to the standard mean and variance of the discretized vector $\bm{f}$. For preprocessing the input (and analogously the output) functions, the statistics are typically computed for each component function and averaged over the training dataset (such that $\mu$ and $\sigma^2$ have as many components as the dimension of the codomain of $f$). Adapting the definition in~\eqref{eq:stats}, it is also straightforward to extend normalization layers, such as batch~\citep{ioffe2015batch}, layer~\citep{lei2016layer}, or instance normalization~\citep{ulyanov2016instance}, to neural operators.

\paragraph{Interpolation}
Interpolation methods, such as spline, kernel, or Fourier interpolations, allow transferring from a discrete to a continuous representation. As mentioned in~\Cref{sec:pointwise}, they can be used after pointwise operators to allow querying at coordinates different from the input discretization.

Interpolation methods can also be used to make standard neural network layers discretization-agnostic. For instance, one can interpolate the input discretization to a latent, fixed discretization, apply a neural network, and then interpolate its output to allow querying at arbitrary points. This defines neural operator layers as in~\Cref{sec:enc_dec}, where the interpolation can be viewed as (non-trainable) encoder and decoder. 

In the same spirit, interpolation methods can also be applied to the input and output of a whole network to define a neural operator~\citep{raonic2024convolutional}. This method is capable of converting any pre-trained neural network into a neural operator. However, this approach suffers from some drawbacks: by fixing a fixed input resolution, the network cannot learn fine-scale details from high-resolution inputs since downsampling the input function leads to a loss of information (e.g., high frequencies for Fourier interpolation). Moreover, the output function space is restricted by the choice of interpolation method, which may dampen the model's expressivity.

\subsection{Summary}
\label{sec: summary conversion}

While there is no very precise and universal automated process to generalize arbitrary neural network layers and architectures to function spaces, there is a consistent underlying strategy that motivates and informs all the examples we have showcased. The first step is to investigate the neural network architecture to understand what would be an analogous transformation in terms of the continuous coordinates of the domain and the functions of interest (i.e., which continuous operator does the neural network roughly discretize). Building on this, the next step is to design strategies to discretize that continuous transformation in a way that respects the core principles presented in \Cref{sec:operator_learning}.

In practice, there are more precise mechanisms following that overall strategy that can be shared by families of neural network layers and architectures. As an example, many neural network architectures can be framed as variants of graph neural networks 
\begin{equation}
     \bm{g}_j = \sum_{i \in \mathrm{Neighb}_j} K_{ij}(\bm{f}_i , \bm{f}_j).
\end{equation}
The common strategy we followed for these neural network architectures is to use the explicit positional information (i.e., the coordinates $x$ on the domain of $f$) and replace the message-passing summation over neighbors by an integral over a suitable neighborhood $D(y_j)$ in the underlying domain:
\begin{equation}
    g(y_j) =  \int_{x \in D(y_j)} \!\!\!\! \K(x,y_j,f(x),f(y_j)) \, \mathrm{d}x.
\end{equation}
This integral operator is analogous to the graph neural network, in terms of the continuous coordinates $x$ and the functions of interest $K$ and $f$. In practice, the integral has to be discretized, which can be achieved by aggregating the contributions in each neighborhood using quadrature weights $\Delta_i$ (see~\Cref{sec:quadrature}), thereby leading to the graph neural operator,
\begin{equation}
  g(y_j) = \sum_{i\colon x_i \in D(y_j)} \!\!\!\! \K(x_i,y_j,f(x_i),f(y_j)) \Delta_i ,
\end{equation}
which respects the core principles of \Cref{sec:operator_learning} and enjoys the desirable properties of neural operators. We have also seen how the exact same strategy allowed to extend multilayer perceptrons to integral operators in \Cref{sec:fnn}, convolutional neural networks to convolutional operators in \Cref{sec:cnn}, graph neural networks to graph neural operators in \Cref{sec:gnn}, and transformer neural networks to transformer neural operators in \Cref{sec:transformer}. In \Cref{fig:layers} we summarize how interpolating discrete convolutional kernels with various methods leads to different ways of constructing $K(x_i, y_j, \cdot, \cdot)$ from $K_{ij}$. For instance, (spherial) FNOs can be derived using global Fourier interpolation.

Other types of layers and operations require different types of modifications so they can be used in a principled manner on function spaces. For instance, encoders and decoders can rely on finite-dimensional latent representations of functions, as long as the encoder is agnostic to the discretization of the input function and the output of the decoder can be queried at arbitrary resolution (see \Cref{sec:enc_dec}). As another example, the key statistics used in normalization layers must account for the functional nature of the data, and this extends to inner products and norms which have to be defined on function spaces instead of vector spaces.

Finally, also note that certain operations are naturally agnostic to the resolution and directly transfer to neural operators, such as operations that are applied pointwise to the discretized function (e.g., activation functions, as well as liftings and projections), as discussed in \Cref{sec:pointwise}.

\section{Training neural operators} \label{sec:training_main}

While the majority of best practices from deep learning and neural network training also directly apply to neural operators (see~\Cref{alg:train}), we present additional considerations that must be taken into account when training neural operators in this section.

\begin{mypython}[alg:train]{Training a neural operator (see~\Cref{sec:training_main})}
neuralop = NeuralOperator(codim_in=codim_in+dim, codim_out=codim_out)
optim = Adam(lr=lr, neuralop.parameters())

def train_iteration(
    f: Tensor[bs, n_in, codim_in], x: Tensor[bs, n_in, dim], delta_x: Tensor[bs, n_in], 
    g: Tensor[bs, n_out, codim_out], y: Tensor[bs, n_out, dim], delta_y: Tensor[bs, n_out]
):
    # Positional encoding and forward pass
    x = 2 ** arange(0, num_frequencies) * pi * x.unsqueeze(-1)
    x = stack([x.sin(), x.cos()], dim=-1)
    f = cat([f, x.view(bs, n_in, -1)], dim=-1)
    g_pred = neuralop(f, x, delta_x, y)
    
    # Loss with quadrature weights and backward pass
    loss = (square(g_pred - g).sum(dim=-1) * delta_y).sum()
    loss.backward()
    optim.step()
\end{mypython}

\subsection{Losses} \label{sec:losses}

For simplicity, we focus on the loss for a single input function and denote by $g$ be the corresponding output of the neural operator provided at discretization points $(y_j)_{i=1}^m$. In practice, we use mini-batches (see~\Cref{sec:training}) and average the losses of multiple input functions.

\paragraph{Data losses} If we have ground truth data $g^\ast$, we can compute \emph{data losses}, but need to make sure that we define the losses on \emph{function spaces}, i.e., they should be agnostic to the resolution. For instance, using a squared $L^2$-loss, we obtain
\begin{equation}
\label{eq:l2}
    \operatorname{Loss}(g,g^\ast) = \sum_{j=1}^m |g(y_j) - g^\ast(y_j) |^2 \Delta_j \approx \int_{D_g} |g(y) - g^\ast(y)|^2 \, \mathrm{d}y,
\end{equation}
where we use quadrature weights $\Delta_j$ for the points $y_j$ to approximate the integral. We observe that for equal quadrature weights $\Delta_i=\Delta$ (e.g., regular grids), the loss in~\eqref{eq:l2} is proportional to the usual mean squared error (MSE).

Analogously, one can also use other norms on function spaces, e.g., \emph{Sobolev norms}, such as the squared $H^1$-loss given by  
\begin{equation}
\label{eq:h1}
    \operatorname{Loss}(g,g^\ast) = \sum_{j=1}^m \Big(|g(y_j) - g^\ast(y_j) |^2 + | \nabla g(y_j) - \nabla g^\ast(y_j) |^2 \Big) \Delta_j.
\end{equation}
By providing supervision on the derivatives of the output functions, Sobolev losses guide the neural operator to accurately capture high-frequency features and smoothness properties of the ground truth function~\citep{li2022learning,kossaifi2023multi}.
Computing Sobolev norms requires the ability to access or approximate the ground truth derivatives. For the output function $g$ of the neural operator, derivatives (w.r.t.\@ to its inputs $y_j$, \emph{not} to the parameters of the neural operator) can be computed using, e.g., \emph{finite-differences}, \emph{Fourier differentiation}, or \emph{automatic differentiation}~\citep{wang2021learning,maust2022fourier,lin2025mGNO}. Note that when the ground truth derivatives cannot be accurately approximated (e.g., when using finite differences at low resolutions), Sobolev losses may be less effective.

\paragraph{Physics-informed losses} If we know that the ground truth operator satisfies a certain (partial) differential equation, (spatiotemporal) derivatives of $g$ can also be used to compute physics-informed losses, similar to \emph{physics-informed neural networks} (\textsc{PINN}s)~\citep{raissi2019physics,sirignano2018dgm}. This can act as a regularizer, reducing the need for extensive training data and enabling the model to be trained with low-resolution or even no data.
In the later case, the PDE loss (together with appropriate boundary and initial conditions) should uniquely describe the solution $g$. 
Instead of relying on a finite number of ground truth input-output pairs, the supervisory signal coming from the PDE loss can be viewed as a \enquote{reward} that can be evaluated at arbitrarily chosen input functions $f$ and corresponding output functions of the neural operator $g$ (discretized at any resolution). While standard \textsc{PINN}s can only solve a single PDE\footnote{Or, more general, families of PDEs parametrized by a \emph{finite-dimensional} parameter.}, physics-informed neural operators can learn the full operator from, e.g., coefficient functions or initial/boundary conditions\footnote{In our notation, the function $f$ collectively describes the considered family of PDEs, e.g., boundary conditions, coefficient functions, and/or geometries (e.g., given in terms of signed distance functions). Since these functions can be defined on different domains, one often uses suitable encodings or extensions to larger domains.} to the solution function~\citep{wang2021learning,li2024physics}. Still, most techniques developed for \textsc{PINN}s (see, e.g.,~\citet{wang2023expert,hao2023pinnacle} for summaries) can directly be extended to physics-informed neural operators.

\paragraph{Balancing multi-objective losses} Scientific machine learning applications often involve predicting multiple variables and minimizing multiple loss terms (e.g., data and physics-informed losses), which are combined into a single objective function. However, appropriate weighting of the different terms is crucial for good performance and well-posedness. Since manual tuning can be a time-consuming and computationally intensive process, structured approaches have been developed that automatically balance losses based on their difficulties, gradient magnitudes, or causal dependencies\footnote{For instance, physics-informed losses have non-unique and often trivial solutions without appropriate boundary conditions (or additional data), motivating to put more weight on the latter in the beginning of the training.}; see, for instance,~\citet{kendall2018multi,lam2023learning,wang2023expert}.

\subsection{Training Strategies}
\label{sec:training}

\paragraph{Data augmentation} For many scientific problems, it is difficult to develop physically-principled data augmentation strategies. We refer to~\citet{brandstetter2022lie} for a discussion on so-called \emph{Lie symmetries} in the context of PDEs.
We can also view different discretizations of input and output functions as a form of data augmentation. This can, for instance, be achieved by subsampling a given, fine discretization. While such \emph{multi-resolution} training can also help neural networks, such as \textsc{CNN}s and \textsc{GNN}s, to work across resolutions, it does not provide any guarantees on their generalization beyond resolutions seen during training. Neural operators, on the other hand, do \emph{not} require multi-resolution training to provide consistent predictions across (unseen) resolutions. However, it can still be employed to accelerate training (e.g., by starting with low-resolution data~\citep{lanthaler2024discretization}) and mitigate discretization errors (using a few samples with higher resolutions). We also note that a certain minimal resolution is required for at least part of the training data to resolve the physics and learn the ground truth operator. Finally, to handle the variable number of discretization points across samples, padding or sparse formats are required for efficient batching, analogous to common practices for \textsc{GNN}s~\citep{pt_geometric}.

\paragraph{Optimization}

As for neural networks, the optimization of neural operators is done using mini-batch stochastic gradient descent with gradients obtained via automatic differentiation, borrowing the same optimizers and learning rate schedulers. In particular, initialization is also crucial to enhance the convergence of stochastic gradient descent~\citep{glorot2010understanding,he2015delving,zhao2021zero,zhang2019fixup}. For both initialization as well as optimization, one also needs to properly deal with complex values arising either in the spectral domain (e.g., for spectral convolutions) or due to complex-valued data (e.g., in quantum applications). This includes complex-valued initializations and arithmetics (in particular, for computing the momentum of optimizers). Finally, since achieving high accuracies is particularly important for scientific applications, we want to highlight research on natural gradients, preconditioning, and quasi-second order optimizers for improved convergence~\citep{muller2024position,goldshlager2024kaczmarz,anil2020scalable}.

\paragraph{Auto-regressive training}
In spatiotemporal forecasting problems, neural operators are often used in an autoregressive fashion, i.e., they are trained to predict a certain time into the future and then iteratively applied to their own output to predict longer time horizons during inference, often referred to as \emph{rollout}. 
However, if the model is only trained on ground truth input data, this can lead to a distribution shift during inference causing instabilities for long autoregressive rollouts~\citep{bonev2023spherical}. To this end, one can add artificial noise during training (see, e.g., \citet{hao2024dpot}), encourage dissipativity~\citep{li2022learning}, or use auto-regressive rollouts also during training. In the latter case, one can avoid excessive memory overheads when backpropagating through the rollout by gradient checkpointing or backpropagating only the last few steps~\citep{lam2023learning,brandstetter2023messagepassingneuralpde}.

\section{Neural Operators in Practice}
\label{sec:nos_in_practice}

\subsection{Applications of Neural Operators}
\label{sec:applications_of_nos}

In \Cref{sec:nns_to_nos}, we demonstrated how layers from popular neural networks can be modified into neural operator layers that remain consistent across resolutions. The properties of neural operators are particularly well-suited for science and engineering applications~\citep{azizzadenesheli2024neural}, and this section highlights a variety of domains where neural operators have already been successfully applied.

\begin{itemize}
    \item \textbf{Fluid and solid mechanics:}  Neural operators have fostered significant advancements in computational mechanics, including modeling porous media, fluid mechanics, and solid mechanics~\citep{you2022learning,choubineh2023fourier}. They offer substantial speedups over traditional numerical solvers while achieving competitive accuracies and expanding their features~\citep{thorsten2023fourcastnet, zhijie2022fourier}. For instance, \textsc{FNO}s constitute the first machine learning-based method to successfully model turbulent flows with zero-shot super-resolution capabilities~\citep{li2020fourier}. \citet{li2022learning} show that Sobolev losses and dissipativity-inducing regularization terms are effective in stabilizing long autoregressive rollouts for highly turbulent flows. Moreover, neural operators have also been used in large eddy simulations of three-dimensional turbulence~\citep{jiang2023efficient} and to learn the stress-strain fields in digital composites~\citep{Rashid2022}. Finally,~\citet{yao2025guided} use neural operators in combination with diffusion models on function spaces to learn distributions over solutions when given sparse or noisy observations.

    \item \textbf{Nuclear fusion and plasma physics:} Neural operators have been used to accelerate magnetohydrodynamic (MHD) simulations for plasma evolution both from state and camera data~\citep{Gopakumar2024,Pamela2025}. Instabilities arising in long-term rollouts using neural operators for plasma evolution have been studied in \cite{Carey2025}, together with the potential of learning across different MHD simulation codes, data fidelities, and subsets of state variables. Furthermore, neural operators have been used for labeling the confinement states of tokamak discharges~\citep{Poels2025}.

    \item \textbf{Geoscience and environmental engineering:} In the geosciences, \textsc{FNO}s and \textsc{UNO}s have been used for seismic wave propagation and inversion~\citep{yang2021seismic,sun2022accelerating}. Extensions of generative models to function spaces have been employed to model earth surface movements in response to volcanic eruptions or earthquakes, or subsidence due to excessive groundwater extraction~\citep{rahman2022generative,seidman2023variational}. Neural operators have also been used to model multiphase flow in porous media, which is critical for applications such as contaminant transport, carbon capture and storage, hydrogen storage, and nuclear waste storage~\citep{gege2022u,Wen2023CO2,Chandra2025}.

    \item \textbf{Weather and climate forecasting:} Versions of \textsc{FNO}s can match the accuracy of physics-based numerical weather prediction systems while being orders-of-magnitude faster~\citep{thorsten2023fourcastnet,mahesh2024huge}. To facilitate stable simulations of atmospheric dynamics on the earth, \cite{bonev2023spherical} introduced the \emph{spherical Fourier neural operator} (\textsc{SFNO}) to extend \textsc{FNO}s to spherical geometries. The super-resolution capabilities of FNOs have also been leveraged for downscaling of climate data, i.e., predicting climate variables at high resolutions from low-resolution simulations~\citep{yang2023fourier}. Additionally, neural operators have been utilized for tipping point forecasting, with potential applications to climate tipping points~\citep{liu2023tipping}.

    \item \textbf{Medicine and healthcare:} Neural operators have been used in multiple settings to improve medical imaging, such as ultrasound computer tomography~\citep{Dai2023,Zeng2023,Wang2025Ultrasound}. As an example, they have been used on radio-frequency data from lung ultrasounds to accurately reconstruct lung aeration maps~\citep{Wang2025Ultrasound}, which can be used for diagnosing and monitoring acute and chronic lung diseases. \citet{jatyani2024unified} use \textsc{FNO}s supplemented with local integral and differential kernels~\citep{liu2024neural} for MRI reconstructions. Neural operators have also been used to improve the design of medical devices, such as catheters with reduced risk of catheter-associated urinary tract infection~\citep{Zhou2024}. Finally, \textsc{GNO}s~\citep{li2020neural,li2020multipole} have been used for spatial transcriptomics data classification~\citep{ahmed2023graph}

    \item \textbf{Computer Vision:} Neural operators have also been applied to several computer vision tasks. For instance, \citet{guibas2021adaptive} use \textsc{FNO}s as token mixers in vision transformer architectures, and \citet{zheng2023fast} use neural operators to accelerate diffusion model sampling. Other works have also explored using \textsc{FNO}s for image classification~\citep{kabri2023resolution} and segmentation~\citep{wong2023fnoseg3d}.
\end{itemize}

\subsection{Empirical Evaluations}
\label{sec:experiments}
In this section, we empirically evaluate different architectures and quantify the difference between using neural networks and neural operators. Specifically, we show that our proposed recipes for converting popular neural network architectures into well-defined operators lead to improved performance and generalization across resolutions.

\begin{figure}[t]
    \centering
    \includegraphics[width=0.95\textwidth]{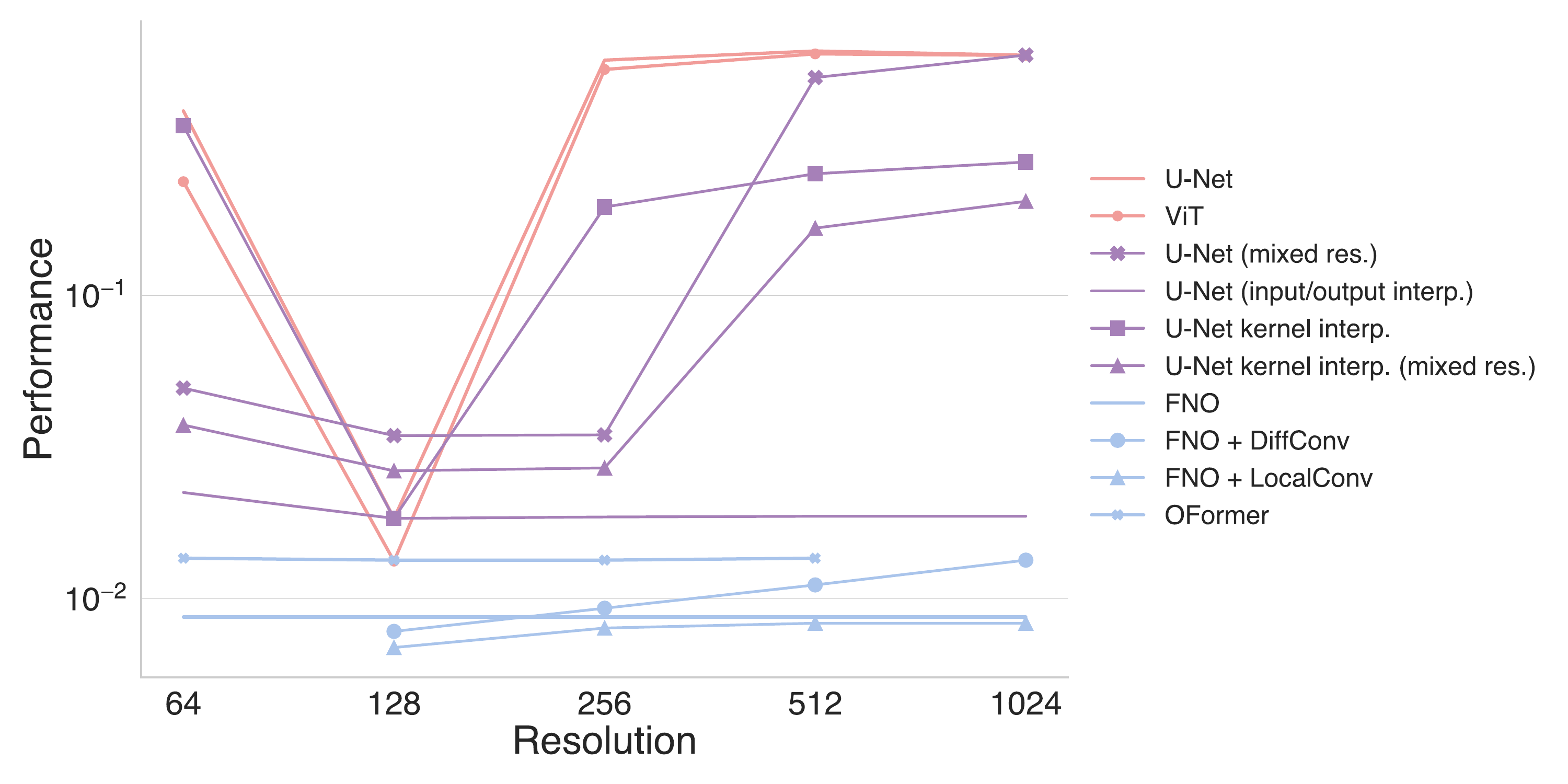}
    \caption{Relative $L^2$-errors (on unseen test data) when training different methods on the Navier-Stokes equations (see~\Cref{sec:experiment_details} for details). Although \textsc{FNO} and \textsc{OFormer} are only trained on resolution $128$, they achieve approximately the same error for higher and lower resolutions. On the other hand, the \textsc{U-Net} only performs well on the training resolution $128$. If we train on mixed resolutions $\{64,128,256\}$ (using each of the $10,\!000$ data points three times with different resolutions in every epoch), it performs well for these resolutions---however, performance still degrades for higher resolutions. Interpolating the convolutional kernels of the \textsc{U-Net} (in both single and mixed-resolution training) also improves generalization across resolutions compared to the baseline \textsc{U-Net}. \textsc{V}i\textsc{T} is included as another common baseline in computer vision, and we observe similar results to those of the \textsc{U-Net}. We optimized the hyperparameters for each method and trained until convergence (see~\Cref{sec:experiment_details}). Due to memory constraints, the \textsc{OFormer} results could not be computed for the highest resolution.}
    \label{fig:empirical_comparison}
\end{figure}

For these experiments, we focus on the well-studied Navier-Stokes~(NS) equations for incompressible fluids, which model the evolution of a fluid acted upon by certain forces. We consider the task of mapping from the force function to the vorticity of the fluid at a later time when initialized at zero vorticity; see \Cref{sec:ns} for details. We compare the performance of different approaches in \Cref{fig:empirical_comparison} and provide details in the following paragraphs.

\paragraph{Receptive field}
When changing the resolution of the input function, neighboring points considered by the kernels of \textsc{CNN}s move closer or farther away from each other; see~\Cref{sec:cnn}. We observe empirically, see \enquote{\textsc{U-Net}} in~\Cref{fig:empirical_comparison}, that such models do not generalize to resolutions different from the training resolution. We observe similar effects when naively applying a \textsc{V}i\textsc{T} (with convolutional tokenization within each patch); see \textsc{V}i\textsc{T} in~\Cref{fig:empirical_comparison}. While the \textsc{V}i\textsc{T} slightly outperforms the \textsc{U-Net}, the performance degrades substantially at resolutions different from the training resolution. The underlying reason is that the receptive field of each layer depends on the resolution. We compared several methods to fix the receptive field:
\begin{itemize}
    \item For instance, one can \emph{linearly interpolate the \textsc{CNN} kernel} (representing the kernel function $K$ in~\Cref{sec:cnn}), improving generalization; see \enquote{\textsc{U-Net} kernel interp.} in~\Cref{fig:empirical_comparison}.
    \item While interpolating the \textsc{CNN} kernel improves generalization, it still suffers from large discretization errors due to the low resolution of the kernel. However, using a \emph{Fourier interpolation via a global spectral convolution} as in the \textsc{FNO} exhibits strong performance and generalization across all considered resolutions.
    \item A global receptive field is also used in \enquote{\textsc{OFormer}}~\citep{li2022transformer}, which relies on a \emph{linear version of self-attention} similar to the one discussed in~\Cref{sec:transformer}. While it suffers from a large computational cost at higher resolutions, it provides good generalization capabilities.
\end{itemize}

\paragraph{Multi-resolution training} Training \textsc{U-Net}s on multiple resolutions, see \enquote{\textsc{U-Net} (mixed res.)} and \enquote{\textsc{U-Net} kernel interp. (mixed res.)} in~\Cref{fig:empirical_comparison}, leads to good test-time performance at these different training resolutions but still poor generalization to unseen resolutions. 
In fact, such augmentation merely \emph{encourage} a model to be agnostic to the discretization observed in the \emph{training} dataset, however do not provide a good inductive bias or theoretical guarantees for unseen resolutions.
Since the training set can only cover a finite set of resolutions in practice, we do not consider neural networks trained with such data augmentation to be a robust and principled method of learning maps between function spaces.

\paragraph{Interpolation} As outlined in~\Cref{sec:aux}, one can also use interpolation, i.e., up- and downsampling, of the input and output and then apply a neural network on the intermediate fixed resolution. While this defines a neural operator (with a fixed encoder-decoder as in~\Cref{sec:enc_dec}) and generalizes across resolutions, see \enquote{\textsc{U-Net} (input/output interp.)} in~\Cref{fig:empirical_comparison}, its performance is limited since it discards any high-resolution information in the input. We note that this loss of high-resolution information could potentially be prevented or mitigated using skip-connections from the high-resolution input and combining it with other neural operator layers.

\paragraph{Composite architectures}
A key principle in architecture design is the idea that different resolution-agnostic layers can be combined to construct architectures that are also resolution-agnostic. In our experiments, we examine two such composite architectures, which we refer to as \textsc{FNO + LocalConv} and \textsc{FNO + DiffConv}, both of which are introduced in \cite{liu2024neural}. \textsc{LocalConv} parametrizes the convolutional kernel $K$ in~\eqref{eq:cnn_fun} as a sum of a finite number of basis functions. \textsc{DiffConv} uses a traditional discrete convolutional kernel (e.g., a $3 \times 3$ kernel) but constrains the parameters to be mean-zero and rescales them according to the resolution. As such, although the receptive field shrinks, the action of the layer provably converges to a directional derivative as the resolution is increased. In \textsc{FNO + LocalConv} and \textsc{FNO + DiffConv}, the \textsc{LocalConv} and \textsc{DiffConv} layers are respectively placed in parallel with the spectral convolution within each layer of the model.

Composite architectures (such as \textsc{FNO + LocalConv} and \textsc{FNO + DiffConv}) inherit the properties of their various components. As such, our results show benefits of combining local (from the \textsc{LocalConv} and \textsc{DiffConv}) and global receptive fields (from the \textsc{FNO}) only at certain resolutions. While all these models are neural operators, different architectures can exhibit varying degrees of discretization error, depending on the training resolution and the data itself. Similarly, every neural operator architecture has a minimal resolution at which this error is controllable and such properties are also inherited by the overall composite architecture.

In summary, our experiments empirically verify three main principles developed in this paper:
\begin{itemize}
    \item To provide the correct inductive bias for cross-resolution generalization, the receptive field of a neural operator must be fixed with respect to the underlying domain of the input function.
    \item Multi-resolution training can improve the performance of neural networks on the training resolutions but not for generalization to unseen resolutions. 
    \item Interpolating the input and output of a neural network to an intermediate fixed resolution defines a naive neural operator but discards high-resolution information in the input.
\end{itemize}

\section{Summary and Outlook}

Successful neural network models, such as \textsc{CNN}s, \textsc{GNN}s, and transformers, have been developed for emulating mappings between finite-dimensional Euclidean spaces. However, for many applications, data arises naturally as functions. Training neural networks on such data requires to discretize the data at a fixed resolutions, constraining their performance, efficiency, and generalization capabilities. In particular, predictions of neural networks made at different resolutions are in general inconsistent with each other.

In this work, we showed how simple modifications can make these popular neural network architectures agnostic to the discretization. This led us to the concept of neural operators as a principled framework for learning mappings between (infinite-dimensional) function spaces. We defined the general design principles and demonstrated how discretization-agnostic versions of popular architectures give rise to general building blocks for neural operators. Finally, we mentioned training recipes and successful applications of neural operators across various applications. 

Since the field of neural operators is still in an early stage, there are a series of open research directions. While the discretization error of neural operators vanishes as the resolution is increased, it is important to be able to quantify and control this error in practical scenarios. The benefits of being discretization-agnostic also sometimes come at the cost of reduced performance at a \emph{fixed} resolution compared to specialized neural network architectures. However, this is not a fundamental shortcoming of neural operators but rather requires further development of architectures with the right inductive biases. As starting points, one can leverage connections of neural operators to successful techniques in computer vision, such as implicit neural representations, or incorporate ideas from classical numerical methods.

Our work provides general guidelines that help practitioners understand the design principles and develop specialized neural operators for their applications. While neural operators have already seen great success in a variety of applications, we believe that they have the potential to accelerate progress and foster breakthroughs across a wide range of additional fields.

\section*{Acknowledgements}
We thank Robert Joseph George for his contributions to the initial drafts. We thank Zijie Li and Md Ashiqur Rahman for fruitful discussions. M. Liu-Schiaffini is supported in part by the Mellon Mays Undergraduate Fellowship. A. Anandkumar is supported in part by Bren endowed chair, ONR (MURI grant N00014-18-12624), and by the AI2050 senior fellow program at Schmidt Sciences.

\newpage 

\bibliographystyle{bibstyle}
\bibliography{bib}
\clearpage 

\appendix

\section{Details}
\label{sec:details}

\subsection{Notation}
\label{sec:notation}

In this section, we provide additional details regarding the notation used in this paper, summarized in~\Cref{tab:notation}.

\paragraph{Continuous functions}

We are interested in learning mappings from a function space $\mathcal{F}$ (\emph{input functions}) to another function space $\mathcal{G}$ (\emph{output functions}). For input functions $f\in \mathcal{F}$, we denote by $D_f$ their domain and by $f(x)$ their evaluation at coordinates $x\in D_f$. Similarly, for output functions $g\in \mathcal{G}$, we denote by $D_g$ their domain and by $g(y)$ their evaluation at coordinates $y\in D_g$. For functions $g=(g_k)_{k=1}^c$ with \emph{codomain} $\R^c$, i.e., $g\colon D_g \to \R^c$, we call the functions $g_k\colon D_g \to \R$ the \emph{component functions} of $g$. While the input and output function spaces can change throughout the layers of a neural operator, we always use the same notation for ease and clarity of presentation.
Moreover, we will typically denote learnable functions of layers by $\K$ (since they correspond to \emph{kernels} of integral operators for many of our considered layers). In practice, these are typically parametrized by neural networks.

\paragraph{Discretized functions} In practice, we will work on discretized functions, and assume that we have access to the function values $f(x_i)$ on a certain point cloud $(x_i)_{i=1}^n$ (see~\Cref{sec:fun_representation} for other representations). Given that the function $f$ has codomain $\R^c$, i.e., $f\colon D_f \to \R^c$, we can represent the function values as a tensor~$\bm{f}$ of shape $n \times c$, where, however, the size $n$ depends on the chosen discretization. Note that we follow the convention of denoting finite-dimensional tensors (including vectors and matrices) by bold variables to distinguish them from functions. Analogously, we assume that the output function $g\colon D_g \to \R^C$ is represented by its evaluations $g(y_j)$ on a point cloud of query points $(y_j)_{j=1}^m$ of interest, leading to a tensor $\bm{g}$ of shape $m \times C$ (where $m$ depends on the number of query points). Finally, for a point $x_i$ in a point cloud, we denote by $\Delta_i$ its \emph{quadrature weight}, specifying its individual weight when discretizing an integral as a weighted sum over the point cloud (see~\Cref{sec:quadrature}). 

If the point clouds correspond to regular grids, we could interpret $x_i$ and $y_j$ as locations of pixel or voxel centers
and $\bm{f}_i$ and $\bm{g}_j$ as the corresponding values. Borrowing terminology from computer vision, this interpretation also motivates to refer to the dimensions $c$ and $C$ of the codomains as the number of input and output \emph{channels}. In the case of regular grids, the quadrature weights are equal, $\Delta_i = \Delta$, and proportional to the reciprocal number of points $\frac{1}{n}$.

\subsection{Function Representations} 
\label{sec:fun_representation}
To perform numerical operations on functions using a computer, suitable discrete representations are required. We focus on the common case where a function $f$ is represented by its point evaluations $f(x_i)$ on a point cloud $x_i\in D_f$. If available, one can additionally consider derivative information, i.e., use 
\begin{equation}
    (x_i,  f(x_i), \nabla f(x_i), \dots).
\end{equation}
In some cases, we might have access to the function $f$ directly, i.e., $f$ is known analytically. This includes cases where $f$ can be specified by coefficients, e.g., w.r.t.\@ a suitable basis or via the parameters of a neural network\footnote{Such neural networks are typically referred to as \emph{neural fields} or \emph{implicit neural representation}.}. This can be viewed as a special case of our setting since we can compute evaluations of $f$ on \emph{arbitrary} points $x$. 
On the other hand, we discuss in \Cref{sec:enc_dec} how to directly operate on coefficients describing a function $f$ (which can be viewed as the output of a given \emph{encoder}). 

\subsection{Neural Operators}
\label{sec:no}
This section provides further details on the definition and properties of neural operators. Mathematically speaking, a \emph{neural operator} is a family of mappings 
\begin{equation}
\label{eq:no_def}
    \textsc{NO}_\theta\colon \mathcal{F} \to \mathcal{G} 
\end{equation}
parametrized by a \emph{finite-dimensional} parameter $\theta\in\R^p$. In the above, $\mathcal{F}$ and $\mathcal{G}$ are suitable function spaces; as an example, we can consider subspaces of continuous functions $ C(\inp{D}, \R^{\inpc})$ and $C(\out{D},\R^{\outc})$, where $\inp{D}\subseteq \R^{\inpd}$ and $\out{D}\subseteq \R^{\outd}$ are suitable domains.
The families $\textsc{NO}_\theta$, $\theta\in\R^p$, are typically designed to be \emph{universal operator approximators}. Specifically, they should be able to approximate any sufficiently regular operator mapping from $\mathcal{F}$ to $\mathcal{G}$ arbitrarily well given a sufficiently large number of parameters $p$. We refer to~\citet{lanthaler2022error,
kovachki2023neural,lanthaler2023nonlocal,lanthaler2023operator,raonic2024convolutional} for corresponding results for some of the neural operators discussed in~\Cref{sec:nns_to_nos}.

In practice, the input functions $f$ are typically represented in the form of a discretization $f|_X =((x_i, f(x_i))_{i=1}^n $ on given points $X= (x_i)_{i=1}^n$; see \Cref{sec:fun_representation}. Since we consider mappings that operate on discretizations with an arbitrary number of points $n$, we introduce the notation
\begin{equation}
    (\inp{D} \times \R^{\inpc})^* \coloneqq \bigcup_{n=1}^\infty (\inp{D} \times \R^{\inpc})^n
\end{equation}    
to denote discretizations of arbitrary size. Our goal is thus to construct \emph{empirical versions}
\begin{equation}
\label{eq:emp_no_def}
    \widehat{\textsc{NO}}_\theta \colon (\inp{D} \times \R^{\inpc})^* \to \mathcal{G}
\end{equation}
of the operator $\textsc{NO}_\theta$ that are \emph{discretization-agnostic}, i.e., that satisfy
\begin{equation}
\label{eq:emp_no_approx}
    \widehat{\textsc{NO}}_\theta \left(  f|_X \right) 
    \approx \textsc{NO}_\theta (f)
\end{equation}
for every $f\in \mathcal{F}$ and discretization\footnote{For some instantiations of neural operators, there can be a lower bound on the resolutions that they can operator. It also depends on the training discretizations and the properties of the ground truth operator whether we can accurately resolve the physics when querying at higher resolutions than the ones seen during training; see \Cref{app:error}.} $X$ of the input domain. 
More precisely, we want to ensure that the error in~\eqref{eq:emp_no_approx} can be made arbitrarily small by considering a sufficiently fine discretization $X$, which can be mathematically formalized using the concept of \emph{discretization convergence}; see \Cref{sec:disc_conv} and~\cite{kovachki2023neural} for details. In particular, the outputs of the empirical version of the neural operator at different discretizations only differ by a \emph{discretization error} that can be made arbitrarily small by increasing the resolution. The following diagram visualizes this concept for the optimal case where we have equality in~\eqref{eq:emp_no_approx} (see~\cite{bartolucci2023neural} for corresponding conditions):
    \begin{center}
    \vspace*{0.5em}
    \begin{tikzcd}[
    column sep=3.5cm, row sep=1cm,
    /tikz/column 1/.append style={anchor=base east, column sep=0pt, inner xsep =0pt}, 
    /tikz/column 3/.append style={column sep=0pt}, 
    /tikz/column 4/.append style={anchor=base west, column sep=0pt, inner xsep =0pt}
    ]
     \mathcal{F} \ni & f \arrow[shorten <= 0.25em,shorten >= 0.25em]{d}[swap]{\text{discretize at }X\ } \arrow[shorten <= 0.5em,shorten >= 0.5em]{r}{\textsc{NO}_\theta}  & g  \arrow[shorten <= 0.25em,shorten >= 0.25em]{d}{\text{\ discretize at }Y}  & \in \mathcal{G} \\
    (\inp{D} \times \R^{c})^*  \ni & f|_X    \arrow[shorten <= 0.5em, shorten >= 0.5em]{ru}[swap]{\widehat{\textsc{NO}}_\theta} \arrow[dashed,swap,shorten <= 0.5em,shorten >= 0.5em]{r}{} & 
    g|_{Y} & \in (\out{D} \times \R^{C})^\ast 
    \end{tikzcd}
    \vspace*{0.5em}
    \end{center}
The dashed line visualizes that, strictly speaking, some neural operator architectures (see, e.g.,~\Cref{sec:pointwise,sec:gnn}) do not output a function but only its evaluation. Specifically, for those architectures, the output of $\widehat{\textsc{NO}}_\theta(f|_X)$ can only be queried on a specific point cloud $Y=(y_i)_{i=1}^m \subset  D_\mathcal{G}$, and not the full domain $D_\mathcal{G}$ since this would require knowledge of $f$ outside of the discretization points $X$. However, for ease of presentation, we assume that the output is suitably interpolated to $D_{\mathcal{G}}$ (see~\Cref{sec:aux}). 
Moreover, we emphasize that the same parameters $\theta$ can be employed across different discretizations of the input and output functions.

In \Cref{app:error}, we demonstrate how such assumptions quantify the \enquote{super-resolution} capabilities of a neural operator, i.e., how sufficiently expressive architectures, trained on a sufficiently fine resolution, are provably approximating the underlying ground truth operator.

\subsection{Discretization Convergence}
\label{sec:disc_conv}

In this section, we formalize the approximation in~\eqref{eq:emp_no_approx} using the concept of discretization convergence. To this end, we need to assume that the discretization is refined in a \enquote{well-behaved} way. In particular, we consider sequences of nested sets 
    \begin{equation}
        X_1 \subseteq X_2 \subseteq \dots \subseteq \inp{D}
    \end{equation} 
    with $|X_n|=n$. We call such a sequence a \emph{discrete refinement} if for any radius $\varepsilon >0$ there exists an index $n$, such that
    \begin{equation}
        \inp{D} \subseteq X_n + B_\varepsilon(0),
    \end{equation}
    i.e., the domain $\inp{D} $ can be covered using balls of radius $\varepsilon$ centered at the $n$ points in $X_n$~\citep{kovachki2023neural}. We now define the discretization convergence of an empirical version $ \widehat{\textsc{NO}}_\theta$ of an operator $ \textsc{NO}_\theta$ as in~\eqref{eq:no_def} and~\eqref{eq:emp_no_def} by the following property: for any discrete refinement $(X_n)_{n\in \N}$ and any compact set $K \subseteq \mathcal{F}$ it holds that
    \begin{equation}
    \label{eq:discr_conv}
        \sup_{f\in K}\, \big\| \widehat{\textsc{NO}}_\theta(f|_{X_n}) - \textsc{NO}_\theta(f)\big\|_{\mathcal{G}} \to 0, \quad n\to \infty, 
    \end{equation}
    where $\|\cdot\|_{\mathcal{G}}$ is a given norm on $\mathcal{G}$; see also~\citet{kovachki2023neural}. 

    In practice, it is important to also consider the convergence speed in~\eqref{eq:discr_conv} since we want the discretization error to be as small as possible for practically relevant numbers of points $n$ and not only in the limit. Moreover, while some layers, e.g., convolutional layers or graph neural network layers with nearest neighbor graphs, are technically discretization convergent, they converge to a typically undesirable pointwise operator; see~\Cref{sec:cnn,sec:gnn}. For the neural operator layers considered in~\eqref{eq:fnn_fun},~\eqref{eq:cnn_fun},~\eqref{eq:gnn_fun}, and~\eqref{eq:transformer_fun}, the discretization convergence depends on the convergence of the utilized numerical integration scheme; see~\Cref{sec:quadrature}.

\subsection{Numerical Integration}
\label{sec:quadrature}

This section provides further details on numerical integration schemes, which can be used to devise empirical versions of integral operator layers that can be applied to discretizations of functions at points $(x_i)_{i=1}^n$. Numerical integration schemes typically take the form 
\begin{equation}
\label{eq:quadrature}
    \int_{D} K(x) \, \mathrm{d}x \approx \sum_{i=1}^n K(x_i) \, \Delta_i
\end{equation}
for quadrature weights $\Delta_i$ and discretization points $x_i$. Different numerical integration schemes (also known as \emph{quadrature} schemes) differ in their choices of weights $\Delta_i$ and quadrature points $x_i$, leading to different errors in the approximation~\eqref{eq:quadrature} for functions $K$ with certain regularity. In particular, this error dictates the discretization error of neural operators defined in terms of integral operators as in~\eqref{eq:fnn_fun},~\eqref{eq:cnn_fun},~\eqref{eq:gnn_fun}, and~\eqref{eq:transformer_fun}.

For Riemann sums, we can partition the domain $D$ into $n$ subdomains $(D_{i})_{i =1}^n$ with $x_i \in D_i$, and choose $\Delta_i = |D_i|$ as the measure of $D_i$. Under mild assumptions on the partition, we obtain convergence to the integral in~\eqref{eq:quadrature} for continuous functions $K$ when $n\to \infty$ and $\max_{i=1}^n\Delta_i \to 0$.

For regular grids, the measure $|D_i|$ corresponds to the product of the grid-spacings in each dimension and is proportional to $\frac{1}{n}$. For general point clouds, one can, for instance, leverage a Delaunay triangulation to obtain a partition;  see~\Cref{alg:quad}. One can also derive methods based on analytically integrating simpler interpolating functions, such as polynomials, e.g., \emph{Newton-Cotes formulas} for regular grids and \emph{Gauss quadrature} for specifically chosen points.

We also want to mention \emph{Monte-Carlo integration} for randomly distributed points $x_i$. To this end, we interpret the integral as an expectation and use a Monte-Carlo estimate, i.e.,
\begin{equation}
\label{eq:mc_int}
\int_{D} K(x) \, \mathrm{d}x =\int_{D}  \frac{K(x)}{p(x)} p(x) \, \mathrm{d}x =  \mathbb{E}_{x\sim p}\left[ \frac{K(x)}{p(x)}\right] \approx \frac{1}{n} \sum_{i=1}^n \frac{K(x_i)}{p(x_i)} =  \sum_{i=1}^n \underbrace{(n p(x_i))^{-1}}_{\coloneqq \Delta_i} K(x_i).
\end{equation}
where the points $x_i$ are assumed to be i.i.d.\@ samples drawn according to a density $p$. 

We note that, e.g., for integral operators, $K$ is learnable, and we can absorb parts of the quadrature weights~$\Delta_i$ that only depend on the point $x_i$ (as opposed to the full set of points or their size $n$) into the kernel. For instance, this includes constant factors. Moreover, for Monte-Carlo integration as in~\eqref{eq:mc_int}, we can also reparametrize $K$ as $\widetilde{K}= Kp$ such that the quadrature weights are $\frac{1}{n}$ for the integrand $\widetilde{K}$. 

\subsection{Fourier Neural Operator}
\label{sec:fno}

This section provides further background information on Fourier neural operators (\textsc{FNO}s)~\citep{li2020fourier}. In their typical form, they first apply a pointwise lifting layer (see~\Cref{sec:pointwise}) to increase the dimension of the codomain (i.e., number of channels) of the input function and, thus, the expressivity\footnote{\citet{lanthaler2023nonlocal} show that only a single Fourier mode, i.e., just a global average, is necessary to achieve universality if the dimension of the codomain is sufficiently large. However, such an approximation is likely inefficient in terms of the required number of parameters.} of the overall architecture. This is followed by a number of \textsc{FNO} blocks, each consisting of a spectral convolution (see~\Cref{sec:cnn}) and a pointwise operator (such as an activation function or an MLP; see~\Cref{sec:pointwise}) combined with skip-connections. The Fourier coefficients are computed separately for each component function, i.e., in practice, the discrete Fourier transform is batched across channels. However, for each frequency, the Fourier coefficients of different component functions are linearly combined, i.e., the channels are mixed in the Fourier space. Since the output of a spectral layer is necessarily bandlimited as the kernel only contains a fixed number of modes, we use a pointwise layer as a skip-connection to retain high frequencies. The last layer of the \textsc{FNO} is given by a pointwise projection layer (see~\Cref{sec:pointwise}) to account for the problem-specific dimension of the codomain of the output function.

As described in~\Cref{sec:enc_dec}, one could replace the Fourier basis in the \textsc{FNO} with other parametrized function classes. However, the former choice has several advantages. First, for regular grids, the \emph{Fast Fourier Transform} (FFT) offers a highly-efficient implementation with reduced computational complexity\footnote{Note that in practice, the discrete Fourier transform can be more efficient than the FFT for sufficiently small numbers of Fourier modes~\citep{lingsch2023beyond}.}. For general grids, the discrete Fourier transforms can be used as an (approximate) mapping between a discretized function and its Fourier coefficients.
Moreover, using a truncated Fourier basis is motivated by the fact that for many PDEs, such as the Navier-Stokes equations, it is known that the spectrum satisfies a certain decay, i.e., the magnitude of the Fourier coefficients decays at a certain rate as the frequency increases. An efficient strategy is thus to progressively increase the number of Fourier modes $k$ together with the input resolution during training~\citep{george2022incremental}. This curriculum learning approach allows the \textsc{FNO} to initially focus on learning low-frequency components of the solution (avoiding wasted computation and overfitting) and progressively adapt to higher-frequency components as the data resolution increases. 

One issue with methods relying on the Fourier basis (as the spectral convolutions in the \textsc{FNO}), is that they make assumptions about the periodicity of the input functions. While non-periodic output functions can still be recovered through skip-connections, non-periodicity of the input function will hinder training by introducing Gibbs phenomena. 
This can be mitigated by periodically extending the input function beyond its original domain, e.g., based on Fourier continuation methods~\citep{maust2022fourier}. Even simple constant padding is often sufficient since the initial lifting layer can mitigate the discontinuities introduced by the padding. Since the input functions are now defined on an extended domain, it is important that the amount of padding changes according to the resolution of the discretization.

\subsection{Attention Mechanism}
\label{sec:attention}

Loosely inspired by human cognition, 
attention mechanisms assign varying levels of importance to different parts of the input, which are often referred to as \emph{tokens}. In our setting, the tokens are given by point evaluations $\bm{f}_i = f(x_i)$ or, after positional encoding, $(x_i,f(x_i))$. Various attention mechanisms have been developed~\citep{bahdanau2014neural,luong2015effective,velivckovic2017graph,vaswani2017attention}. For instance, the popular \emph{soft(arg)max with dot-product attention}  is given by
\begin{equation}
\label{eq:self-attention}
   \bm{g}_j  = \sum_{i=1}^n  \frac{\exp( \tau \langle k(\bm{f}_i), q(\bm{f}_j) \rangle)}{\sum_{\ell=1}^n \exp( \tau \langle k(\bm{f}_\ell), q(\bm{f}_j) \rangle)} v(\bm{f}_i),
\end{equation}
where $v,$ $k$, and $q$ are learnable \emph{query}, \emph{key}, and \emph{value} functions. The inner product in~\eqref{eq:self-attention} is rescaled by a so-called temperature $\tau$, often chosen as $\tau =(d_{\mathrm{att}})^{-1/2}$, where $d_{\mathrm{att}}$ is the dimension of the codomain of $k$ and $q$. Moreover, multiple attention mechanisms, so-called \emph{heads}, can be combined~\citep{vaswani2017attention}. 

We speak of \emph{self-attention} when evaluating the query and key functions at the same tokens (as in~\eqref{eq:self-attention}). In the more general case of \emph{cross-attention}, we evaluate the query function $q$ in~\eqref{eq:self-attention} on different tokens. In our neural operator layer in~\eqref{eq:transformer_fun}, this can be another function $\tilde{f}$, potentially discretized on a different point cloud $(\tilde{x}_i)_{i=1}^{\tilde{n}}$. Since the point cloud $\tilde{x}_i$ defines the coordinates on which we can query the output function $g$ of the attention layer, we can also use cross-attention to evaluate the layer on arbitrary coordinates by picking $\tilde{f}$ to be a prescribed function, e.g., the identity or a positional embedding.

\subsection{Special Cases of Encoders \& Decoders}
\label{sec:special_enc_dec}

In this section, we show that certain integral operators (see~\Cref{sec:gnn}), including spectral convolutions (see~\Cref{sec:cnn}), can be viewed as special cases of encoder-decoder layers introduced in~\Cref{sec:enc_dec}. Specifically, we consider encoder-decoder layer of the form
\begin{equation}
\label{eq:special_enc_dec}
    \bm{v}_j = \int_{D_f} [b_j(x)]^* f(x)   \, \mathrm{d}x, 
    \quad \bm{w}_j = \bm{K}_j \bm{v}_j,
    \quad \text{and} \quad g(y) = \sum_{j=1}^{k} \beta_j(y) \bm{w}_j,
\end{equation}
with input function $f\colon D_f \to \R^c$ and output function $g\colon D_g \to \R^C$.
This can be interpreted as an encoder based on inner products with functions $b_j$, linear maps $\bm{K}_j$ in the latent space, and a decoder based on linear combinations with functions $\beta_j$. Thus, the overall layer can be expressed as
\begin{equation}
\label{eq:enc_dec_gno}
    g(y) = \int_{D_f}  \underbrace{\sum_{j=1}^{k} \beta_j(y) \bm{K}_j [b_j(x)]^*}_{ \coloneqq K(x,y)} f(x)  \, \mathrm{d}x,
\end{equation}
which corresponds to a version of the integral operator in~\eqref{eq:fnn_fun} and~\eqref{eq:gnn_fun}. 
We also recover the following special cases:
\begin{itemize}
    \item We obtain a layer of the \emph{low-rank neural operator}~\citep{kovachki2023neural,lanthaler2023nonlocal} when setting 
    $\bm{K}_j=1$ and learning $b_j\colon D_f \to \R^c$ and $\beta_j\colon D_g\to \R^C$.
    \item The spectral convolution in~\Cref{sec:cnn} is given by setting $b_{j+1}(x)=\beta_{j+1}(x)=e^{2\pi i j x}$, i.e., using the first Fourier basis functions, and learning $\bm{K}_j \in \mathbb{C}^{C \times c}$, where we assume that the input function is defined on the torus.
\end{itemize}

\section{Error Analysis for Zero-Shot Predictions on Different Discretizations}
\label{app:error}

In this section we provide a high-level error analysis to outline the interplay between different errors, i.e., discretization, optimization, and approximation errors. For a detailed analysis and corresponding bounds, we refer, e.g., to~\citet{kovachki2021universal,lanthaler2022error,de2022generic,lanthaler2023nonlocal,kovachki2023neural} for approximation and generalization results, to~\citet{kovachki2024data,grohs2025theory} for results on sampling complexities, and to~\citet{lanthaler2024discretization} for results on the discretization error.

Here, we consider the use of neural operators to predict on discretizations different than the training discretization, without retraining. In particular, this includes zero-shot super-resolution where the query discretization has a higher resolution than the training discretization. To this effect, let

\begin{itemize}\setlength{\itemindent}{4mm}
\item $\textsc{NO}^\ast\colon\mathcal{F}\to \mathcal{G}$ be the ground truth operator between Banach spaces $\mathcal{G}$ and $\mathcal{F}$,
    \item $\textsc{NO}_\theta\colon\mathcal{F}\to \mathcal{G}$ be a neural operator with parameter set $\theta \in \Theta$,   
    \item $\widehat{\textsc{NO}}_\theta$ denote an empirical version of the neural operator $\textsc{NO}_\theta$ as in~\eqref{eq:emp_no_def},
    \item $X$ and $\widetilde{X}$ be fixed discretizations of the domain (the training and query discretizations, respectively). 
\end{itemize} 
Suppose we trained the neural operator on functions in $\mathcal{F}$ discretized on $X$, and obtained learned parameters $\hat{\theta} \in \Theta$. We are are interested in bounding the error of the neural operator on a function $f\in \mathcal{F}$ discretized on the (unseen) query discretization $\widetilde{X}$, i.e., 
\begin{equation}
\label{eq:no_error}
        \eps(f,\widetilde{X}) \coloneqq  \| \textsc{NO}^*(f) - \widehat{\textsc{NO}}_{\hat{\theta}}(f|_{\widetilde{X}})\|_{\mathcal{G}}.
        \end{equation}
Using the triangle inequality, we obtain
\begin{align}
\label{eq:no_error_bound}
     \eps(f,\widetilde{X}) \le  \underbrace{ \| \textsc{NO}^*(f) - \widehat{\textsc{NO}}_{\hat{\theta}}(f|_{X})\|_{\mathcal{G}}}_{= \, \vphantom{\widetilde{X}} \eps(f,X)}  + \underbrace{ \| \widehat{\textsc{NO}}_{\hat{\theta}}(f|_{X}) - \textsc{NO}_{\hat\theta}(f) \|_{\mathcal{G}}}_{\le \, \vphantom{\widetilde{X}} \eps_{\mathrm{discr}}(f,X) } +   \underbrace{ \| \textsc{NO}_{\hat{\theta}}(f)- \widehat{\textsc{NO}}_{\hat{\theta}}(f|_{\widetilde{X}})\|_{\mathcal{G}}}_{\le \, \eps_{\mathrm{discr}}(f,\widetilde{X})},
\end{align}
where $\eps(f,X)$ is the error of the neural operator on the discretization $X$ defined as in~\eqref{eq:no_error}, $\eps_{\mathrm{discr}}(f,X)$ is a bound on the \emph{discretization error} of $\textsc{NO}$ on $X$, i.e.,
\begin{equation}
\label{eq:discr_error}
     \sup_{\theta\in \Theta} \| \textsc{NO}_\theta(f) - \widehat{\textsc{NO}}_\theta(f|_{X})\|_{\mathcal{G}} \le \eps_{\mathrm{discr}}(f,X),
\end{equation}
and $\eps_{\mathrm{discr}}(f,\widetilde{X})$ is defined analogously. Under suitable conditions for discretization convergence (see \Cref{sec:disc_conv}), the discretization errors $\eps_{\mathrm{discr}}$ go to zero as the resolutions of the discretizations increase. In this case, for high resolutions, the inequality in~\eqref{eq:no_error_bound} shows that the error for the discretization $\widetilde{X}$ is essentially bounded by the error for the discretization $X$.

Next, we want to bound the error on the training discretization $\eps(f,X)$ in~\eqref{eq:no_error_bound}. To this end, let $\theta^\ast\in\Theta$ be a parameter that achieves minimal \emph{approximation error} to the ground truth operator, i.e.,
\begin{equation}
\label{eq:approx_error}
   \sup_{f\in \mathcal{F}} \| \textsc{NO}^\ast(f) - \textsc{NO}_{\theta^\ast}(f) \|_{\mathcal{G}}  = \infp_{\theta \in \Theta} \, \sup_{f\in \mathcal{F}} \|  \textsc{NO}^\ast(f) - \textsc{NO}_{\theta}(f) \|_{\mathcal{G}} = \eps_{\mathrm{approx}}.
\end{equation} 
Moreover, suppose the \emph{optimization error}\footnote{For simplicity, we do not explicitly consider the \emph{generalization error} arising from training only on a finite number of samples.} between the trained and optimal neural operator can be bounded by some $\eps_{\mathrm{opt}}$, i.e.
\begin{equation}
\label{eq:opt_error}
\sup_{f\in \mathcal{F}} \| \textsc{NO}_{\theta^\ast}(f)  - \textsc{NO}_{\hat{\theta}}(f) \|_{\mathcal{G}} \le \eps_{\mathrm{opt}}.\end{equation}
Then we can use the triangle inequality to bound the model error on the training discretization as
\begin{align}  
\eps(f,X)  \le   \underbrace{\| \textsc{NO}^*(f) - \textsc{NO}_{\theta^*}(f) \|_{\mathcal{G}}}_{\le \, \eps_{\mathrm{approx}}}  +  \underbrace{ \| \textsc{NO}_{\theta^*}(f)  - \textsc{NO}_{\hat{\theta}}(f) \|_{\mathcal{G}}}_{\le \, \eps_{\mathrm{opt}}} 
+  \underbrace{\| \textsc{NO}_{\hat \theta}(f) -\widehat{\textsc{NO}}_{\hat \theta}(f|_{X})\|_{\mathcal{G}}}_{ \le \, \eps_{\mathrm{discr}}}.
\end{align}
In particular, substituting this inequality in \eqref{eq:no_error_bound}, implies that
\begin{equation} \label{eq:no_error_bound_2}
         \eps(f,\widetilde{X})  \le  \eps_{\mathrm{approx}} + \eps_{\mathrm{opt}}  +   2 \eps_{\mathrm{discr}}(f,X) + \eps_{\mathrm{discr}}(f,\widetilde{X}). 
\end{equation}
This shows that we can bound the error $\eps(f,\widetilde{X})$ for a function discretized on $\widetilde{X}$, by the approximation error in~\eqref{eq:approx_error}, the optimization error in~\eqref{eq:opt_error}, and the discretization errors for $X$ and $\widetilde{X}$ as in~\eqref{eq:discr_error}. Due to the resolution convergence of the neural operator, the discretization errors vanish as the resolution increases. The prediction error on discretizations with higher resolution than the training data can thus be made arbitrarily small by having a \emph{sufficiently expressive} neural operator that is trained on a \emph{sufficiently high resolution} to \emph{near-optimality}. In particular, this provides a theoretical explanation for the super-resolution capabilities of neural operators that approximate the underlying ground truth operator.

\section{Experimental Details}
\label{sec:experiment_details}

\begin{figure}[t]
    \centering
    \includegraphics[height=7cm]{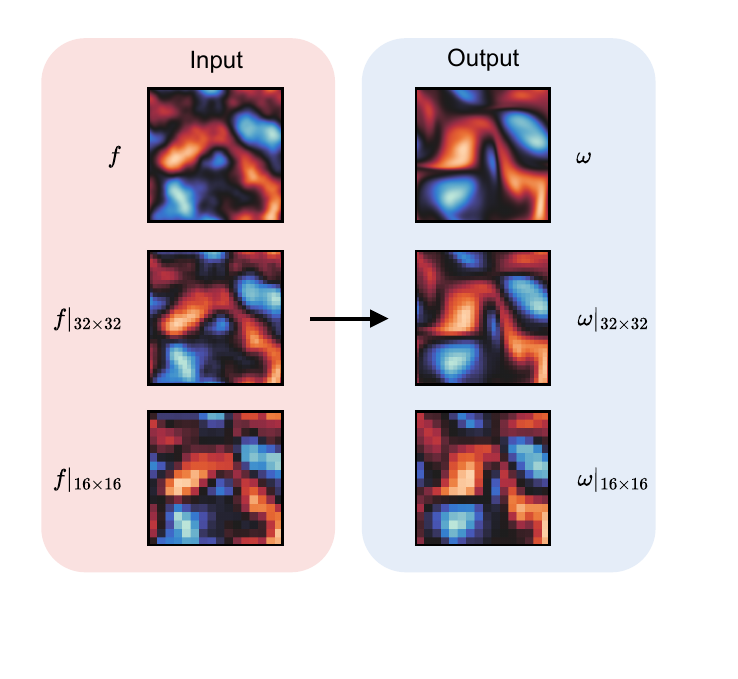}
    \caption{Sample input and output functions and two discretizations for learning the mapping from forcing function to vorticity governed by the Navier-Stokes equations in~\Cref{sec:ns}.}
    \label{fig:ns_discretizations}
\end{figure}

In this section, we describe the dataset as well as implementation and hyperparameter details for our experiments. As mentioned in \Cref{sec:training}, the general training procedure for neural operators is similar to that of neural networks (see \Cref{alg:train}).

\subsection{Navier-Stokes Equations}
\label{sec:ns}

Since the primary applications of neural operators have been in learning the solution operators for PDEs, the experiments in this paper focus on one well-studied PDE: the Navier-Stokes (NS) equations for incompressible fluids. As in~\cite{kossaifi2023multi}, we consider the vorticity-stream ($\omega$ -- $\psi$) formulation of the $2$-dimensional NS equations, given by
\begin{equation}
\begin{aligned}
    \partial_t \omega + \nabla^\perp \psi \cdot \omega = \frac{1}{\textrm{Re}} \Delta \omega + f, \hspace{1cm} &\text{on} \ \ \mathbb{T}^2 \times (0,T] \\
    -\Delta \psi = \omega, \quad \int_{\mathbb{T}^2} \psi = 0 \hspace{1cm} &\text{on} \ \ \mathbb{T}^2 \times (0,T].
\end{aligned}
\end{equation}
In the above, $\omega$ and $\psi$ denote the vorticity and stream functions defined on the torus $\mathbb{T}^2$ and the time interval $[0,T]$, $f$ is a forcing function distributed according to a Gaussian measure, and $\textrm{Re} > 0$ is the Reynolds number, which can be used to characterize the laminar or turbulent nature of the flow. As in~\cite{kossaifi2023multi}, we fix the initial vorticity to be $\omega(\cdot,0) = 0$ and seek to learn the nonlinear operator $f \mapsto \omega(\cdot,T)$ from the forcing function to the vorticity at some fixed time $T > 0$; see also \Cref{fig:ns_discretizations}.
Following \cite{kossaifi2023multi}, we use $10,000$ samples for training and $2,000$ samples for testing.

\subsection{Implementation Details}

In order to maintain the fairness of comparisons, we attempted to maintain an equal number of parameters (around $63$M) across all models throughout our experiments (if allowed by GPU memory constraints). Furthermore, all models were trained until convergence using the Adam optimizer~\citep{kingma2014adam} with a weight decay of $10^{-4}$, except the \textsc{OFormer} which used a weight decay of $10^{-6}$. Training samples are given on a uniform spatial discretization of resolution $128 \times 128$. Where GPU memory allowed, we evaluated our models at several data resolutions, as shown in~\Cref{fig:empirical_comparison}. All models are trained on absolute $L^2$-error and evaluated using relative $L^2$-error. When available, we use the implementations found in the \textsc{NeuralOperator} Python library\footnote{The \textsc{NeuralOperator} library is publicly available at \href{https://github.com/neuraloperator/neuraloperator}{\nolinkurl{github.com/neuraloperator/neuraloperator}}. Our code is available at \href{https://github.com/neuraloperator/NNs-to-NOs}{\nolinkurl{github.com/neuraloperator/NNs-to-NOs}}.}~\citep{kossaifi2024neural}. All models were trained on a single NVIDIA RTX 4090 GPU, with the exception of the \textsc{OFormer}, which was trained on a single NVIDIA H100 GPU due to higher memory requirements.

Note that for some models, multiple resolutions were used during training. In these cases, we use each of the $10,000$ samples at each resolution in each epoch. The specific training resolutions used for these particular models are described below, along with the implementation details for all models used in our experiments.

\paragraph{\textsc{U-Net}} Our \textsc{U-Net} model is adopted from the \textsc{PDEArena}~\citep{pdearena} baseline\footnote{The \textsc{PDEArena} code is publicly available at \href{https://github.com/pdearena/pdearena}{\nolinkurl{github.com/pdearena/pdearena}}.}. In our experiments, we set the number of hidden channels to be $19$, the kernel size to be $9$, and use layer normalization. We use channel multipliers of $1$, $2$, $2$, $3$ for each resolution; see \cite{pdearena} and the official repository for more details. We halve the learning rate every $50$ epochs, starting at an initial learning rate of $5 \cdot 10^{-4}$.

\paragraph{\textsc{U-Net} with mixed resolution training}
In these experiments, we use the same \textsc{U-Net} architecture as above; however, we augment the training data by including the samples at three different resolutions. In addition to resolution $128 \times 128$, we supplement the training set with the same data at resolutions $64 \times 64$ and $256 \times 256$. We halve the learning rate every $20$ epochs, starting at an initial learning rate of $5 \cdot 10^{-4}$.

\paragraph{\textsc{U-Net} with input \& output interpolation}
\textsc{U-Net} with input and output interpolation fixes the \textsc{U-Net} resolution at $128 \times 128$ and interpolates the input and output to the desired resolution, as described in~\Cref{sec:aux}. For these experiments, we use the same \textsc{U-Net} model as described above. We use bilinear interpolation in our experiments.

\paragraph{FNO} In our experiments, we set the number of Fourier modes to $64$, the number of hidden channels to $62$, and we use $4$ Fourier layers consisting of spectral convolutions, pointwise MLPs, and soft-gating skip-connections as proposed by~\cite{kossaifi2023multi}. We halve the learning rate every $33$ epochs, starting at an initial learning rate of $5 \cdot 10^{-4}$.

\paragraph{Local neural operator}
In our experiments, we follow \cite{liu2024neural} in supplementing each spectral convolution with either local integral or differential operators. For both experiments, we use $40$ Fourier modes and $99$ channels, such that the total number of parameters is similar to the \textsc{FNO} experiments. Decreasing the number of Fourier modes encourages the \textsc{FNO} to learn global features, and increasing the number of channels increases the expressivity of the local operators for capturing high-frequency features. Apart from these changes, we follow the same settings as for the \textsc{FNO} architecture described above. For \textsc{FNO + DiffConv}, we only supplement the last layer with a differential layer, whereas for \textsc{FNO + LocalConv}, we supplement all four layers with a local integral layer. For \textsc{FNO + LocalConv}, we use the same basis as in \cite{liu2024neural} with a radius cutoff of $0.03125$.

\paragraph{Vision transformer}
For the vision transformer (\textsc{V}i\textsc{T}) experiments, we adapt the model from \cite{vit}, using the implementation in the \textsc{HuggingFace Transformers} library\footnote{The \textsc{Transformers} library is publicly available at \href{https://github.com/huggingface/transformers}{\nolinkurl{github.com/huggingface/transformers}}.}. We use a patch size of $16 \times 16$, a hidden size of $768$, $12$ hidden layers, $12$ attention heads, an intermediate size of $1900$, and GeLU activations. We halve the learning rate every $50$ epochs, starting at an initial learning rate of $5 \cdot 10^{-4}$.

\paragraph{\textsc{OFormer}}
In our experiments, we adapt the \textsc{OFormer} architecture from~\cite{li2022transformer}. However, we introduce two modifications that helped for high-resolution uniform grids:
\begin{enumerate}
    \item Since the input and output grids are the same, we removed the decoder, resulting in an encoder-only architecture.
    \item To better capture low frequencies within each encoder layer, we summed the output of the attention block and the residual connection with the output of a spectral convolution~\citep{li2020fourier} applied on the input, normalized by $\sqrt{3}$ for stability.
\end{enumerate}
These two modifications significantly improved the results over the baseline \textsc{OFormer} on this task. We use $220$ hidden channels and $16$ attention heads in the encoder. We used $4$ encoder layers and $16$ modes in the spectral convolutions. We halved the learning rate every $33$ epochs, with an initial learning rate of $10^{-4}$.

\end{document}